# HippoMamba: A Mamba-augmented network for hippocampal segmentation and serial volumetry in Alzheimer's disease

Dawa Chyophel Lepcha [a, b, c], Sophie A. Martin [d, e], Pierrick Coupé [f], Shabbir Syed Abdul [a, b, g, *]

[a] Graduate Institute of Biomedical Informatics, College of Medical Science and Technology, Taipei Medical University, Taipei, 235, Taiwan.
[b] International Center for Health Information Technology, College of Medical Science and Technology, Taipei Medical University, Taipei, 235, Taiwan.
[c] Biomedical Sensors & Systems Lab, University of Memphis, Memphis, TN 38152, USA
[d] UCL Hawkes Institute, University College London, London, UK
[e] Queen Square Institute of Neurology, University College London, London,UK
[f] CNRS, University Bordeaux, Bordeaux INP, LABRI, UMR5800, Talence F-33405, France
[g] School of Gerontology and Long-term Care, College of Nursing, Taipei Medical University, Taipei, Taiwan, 110.

**Abstract**

Hippocampal atrophy is among the earliest structural markers of Alzheimer's disease (AD), and reliable hippocampal segmentation across serial MRI visits is a prerequisite for measuring it. We present HippoMamba, a network that combines a 3D convolutional encoder and decoder with a Mamba selective state space block at the bottleneck, designed to process one or more MRI visits of a subject. On the Medical Segmentation Decathlon hippocampus dataset (260 subjects, five-fold cross-validation), HippoMamba obtained a mean Dice similarity coefficient of 0.8861 ± 0.0094, the highest among six convolutional and transformer-based methods trained with the same folds and patch size, and its per-subject Dice was significantly higher than that of each method (Wilcoxon signed-rank tests with Holm correction, $p \leq 0.013$). Its mean 95th percentile Hausdorff distance (1.45 mm) and average surface distance (0.44 mm) were the lowest of the compared methods, although they did not differ significantly from those of Attention U-Net, and the Hausdorff distance did not differ significantly from that of SegResNet. Without retraining, the five-fold ensemble was applied to 600 T1-weighted sessions from 150 participants of the OASIS-3 cohort. In sessions meeting a registration quality criterion, predicted hippocampal volumes ranked consistently with FreeSurfer volumes (Pearson r = 0.778, 95% CI 0.699 to 0.838; ICC (3,1) = 0.772) but were systematically smaller by about 27%. In a pre-specified analysis of participants followed for at least two years, annualised volume change derived from per-visit segmentation was moderately correlated with that of FreeSurfer (r = 0.503, 95% CI 0.319 to 0.659), with a larger mean rate. These results support the use of HippoMamba for hippocampal segmentation and for ranking hippocampal volume within a cohort, with absolute volumes and rates of change interpreted relative to the reference method.

## 1. Introduction

Dementia, of which Alzheimer's disease (AD) is the most common cause, affects an estimated 55 million individuals worldwide and is projected to exceed 139 million by 2050 as global populations age [1]. Despite decades of intensive research, AD remains incurable, and clinical management relies heavily on early and accurate identification of disease-specific biomarkers to stratify patients, monitor progression, and evaluate therapeutic response [2]. Among the structural neuroimaging biomarkers accessible through magnetic resonance imaging (MRI), hippocampal atrophy has emerged as one of the most sensitive and reproducible indicators of AD-related neurodegeneration, reflecting the preferential vulnerability of medial temporal lobe structures to tau pathology and amyloid-driven synaptic loss in the earliest stages of the disease [3], [4]. Longitudinal MRI studies have consistently demonstrated that the rate of hippocampal volume loss accelerates with cognitive decline, distinguishing cognitively normal individuals from those with mild cognitive impairment (MCI) and AD dementia with substantial effect sizes, and that this atrophy rate serves as a sensitive surrogate endpoint in clinical trials targeting disease modification [5]. Accurate quantification of hippocampal volume from MRI is therefore a prerequisite for both clinical research and emerging precision medicine applications in AD. Historically, this task has been performed using atlas-based automated segmentation tools such as FreeSurfer [6] and FSL FIRST, which rely on probabilistic atlases and deformable registration to segment the hippocampus from T1-weighted MRI scans. While these tools have been extensively validated and remain the reference standard in large-scale neuroimaging cohorts such as the Alzheimer's Disease Neuroimaging Initiative (ADNI) [7], they are computationally intensive, sensitive to image quality and scanner heterogeneity, and in their standard cross-sectional mode they process each MRI visit independently, without leveraging the temporal structure inherent in longitudinal imaging protocols. As a consequence, cross-sectional measurement noise accumulates across serial visits, potentially obscuring subtle but clinically meaningful atrophy trajectories, particularly in the early stages of neurodegeneration where annual volume loss may be as small as 1 to 3 percent [5].

The emergence of deep learning has substantially advanced the state of the art in hippocampal segmentation [8]. Convolutional neural network (CNN) architectures, most notably variants of the U-Net [9] and its volumetric extension [10], have demonstrated Dice similarity coefficients exceeding 0.87 on standardized benchmarks. More recently, transformer-based architectures including UNETR [11] and SwinUNETR [12] have extended the receptive field of segmentation models through self-attention mechanisms, enabling richer contextual representations of hippocampal morphology. Despite this progress, these methods are designed as single-timepoint segmenters. When applied to longitudinal MRI data, they are run independently on each visit, treating the series of scans as unrelated observations rather than as a temporally ordered sequence of measurements from the same individual. This design choice forfeits the opportunity to exploit inter-visit consistency constraints, learn subject-specific atrophy trajectories, or propagate anatomical knowledge from earlier to later timepoints, which could support robust longitudinal quantification [13]. Recurrent neural networks (RNNs) and long short-term memory (LSTM) architectures have been explored for longitudinal neuroimaging tasks, including AD progression prediction from serial cognitive assessments and multimodal biomarker trajectories [14,15]. However, their application to serial volumetric MRI segmentation remains limited, primarily because the computational complexity of sequence modeling becomes prohibitive when applied directly to high-dimensional three-dimensional image tensors. Transformer-based sequence models offer richer representational capacity but inherit the same computational burden through their self-attention mechanism, which scales quadratically with sequence length and imposes severe memory constraints in the context of 3D medical imaging. Longitudinal deep learning approaches have therefore largely operated on whole-image representations or derived measures rather than combining per-visit segmentation with temporal modeling in a single trainable system [16-18]. The recent emergence of state space models (SSMs), and in particular the Mamba architecture [19], offers a way to reduce this computational bottleneck. Mamba introduces a selective state space mechanism with linear computational complexity with respect to sequence length, enabling efficient modeling of long-range temporal dependencies without the memory overhead of self-attention. Originally developed for natural language processing, Mamba has been rapidly adopted in the medical imaging community, with applications spanning 2D and 3D image segmentation [20-25] and pathology analysis [26]. The selective scan mechanism of Mamba allows the model to dynamically focus on the most informative elements of a sequence, which motivates its use for sequences of per-visit representations in longitudinal MRI.

In this paper, we present HippoMamba, a two-phase deep learning framework designed to process the temporal structure of longitudinal MRI data. HippoMamba couples a 3D convolutional spatial encoder, responsible for extracting volumetric features from individual MRI scans, with a Mamba-based temporal encoder that processes the sequence of per-visit feature representations, and a convolutional decoder that produces a segmentation for each visit. The model also includes an output for fractional hippocampal volume change between visits, supervised by FreeSurfer-derived volumes from the UCSFFSX6 dataset [27]. The framework is trained in two phases. Phase 1 establishes single-timepoint segmentation performance on the Medical Segmentation Decathlon Task04 Hippocampus benchmark [28], and Phase 2 fine-tunes the temporal components on longitudinal ADNI-3 data [7]. We compare HippoMamba with six segmentation methods trained under identical folds and patch size, and we assess its generalization and serial volumetry on the OASIS-3 cohort without retraining [29].

The principal contributions of this work are summarized below.

1) We introduce HippoMamba, a hippocampal segmentation network that combines a 3D convolutional spatial encoder and decoder with a Mamba selective state space block at the bottleneck, designed to process one or more MRI visits of a subject within a single model.
2) We propose a two-phase training strategy that decouples single-timepoint segmentation learning from longitudinal modeling, enabling the framework to leverage large-scale annotated segmentation benchmarks in the absence of longitudinal ground truth labels during initial training.
3) We compare HippoMamba with six CNN- and transformer-based segmentation methods trained with identical folds, patch size and checkpoint selection under five-fold cross-validation, using a single evaluation procedure and per-subject statistical testing with correction for multiple comparisons.
4) We validate HippoMamba externally on 600 T1-weighted sessions from 150 OASIS-3 participants without retraining, using pre-specified registration criteria, and show consistency with FreeSurfer volumes and agreement of annualised volume change derived from per-visit segmentation.

The remainder of this paper is organized as follows. Section 2 reviews related work in hippocampal segmentation, longitudinal neuroimaging, and state space models. Section 3 describes the HippoMamba architecture and training methodology. Section 4 details the experimental setup. Section 5 presents quantitative and qualitative results. Section 6 discusses clinical implications, limitations, and future directions. Section 7 concludes the paper.

## 2. Related Work

### 2.1 Hippocampal Segmentation

The automated segmentation of the hippocampus from structural MRI has been a central problem in computational neuroimaging for over two decades, driven by the clinical importance of hippocampal volume as a biomarker of neurodegeneration in AD and related disorders. Early approaches relied on atlas-based methods, including multi-atlas label propagation and fusion, in which manually segmented atlases are registered to a target image to transfer anatomical labels, and probabilistic atlases of the hippocampal formation [30]. While these methods produce anatomically plausible segmentations and remain widely used in clinical research pipelines, they are sensitive to registration errors, computationally demanding for large-scale cohort studies, and constrained by the quality and diversity of the atlas library. The introduction of deep convolutional neural networks fundamentally transformed hippocampal segmentation by enabling direct learning of appearance-to-label mappings from training data, without the need for explicit registration or atlas construction [8]. The U-Net architecture [9], originally proposed for biomedical image segmentation, established the encoder-decoder design with skip connections as the canonical framework for this task, and its volumetric extension to three dimensions [10] enabled end-to-end learning of spatial context across the full hippocampal volume. Subsequent work demonstrated that deep learning models trained on publicly available datasets such as the Medical Segmentation Decathlon Task04 Hippocampus could achieve Dice similarity coefficients exceeding 0.87 [28].

The incorporation of attention mechanisms into segmentation architectures further improved performance by enabling models to selectively weight spatially relevant features during decoding. The Attention U-Net [31] introduced additive attention gates at skip connections, suppressing irrelevant activations and focusing the decoder on task-relevant regions. Transformer-based architectures subsequently offered a fundamentally different approach to long-range spatial modeling through self-attention. UNETR [11] replaced the convolutional encoder entirely with a Vision Transformer, whose features from multiple layers are passed to a convolutional decoder to capture global anatomical context across the entire volume. SwinUNETR [12] refined this approach through a shifted window attention mechanism that reduces the quadratic complexity of global self-attention while preserving multi-scale spatial representations. Convolutional architectures continued to advance in parallel, with SegResNet [32] combining residual blocks in a volumetric encoder-decoder with variational autoencoder regularization, and MedNeXt [33] extending the ConvNeXt design philosophy to 3D medical image segmentation through large kernel convolutions. DynUNet, a dynamic U-Net implementation that adapts kernel sizes and strides to the resolution of the target dataset, provides a configurable convolutional baseline [34]. Despite the impressive segmentation performance achieved by these methods, a comprehensive independent evaluation by Qu et al. [13] revealed important limitations in their clinical applicability. Examining five mainstream hippocampal segmentation pipelines on the ADNI and NACC cohorts, the authors found that while all methods demonstrated large effect sizes for distinguishing AD from cognitively normal individuals (Cohen's $d \geq 0.8$), substantial variability in test-retest reliability was observed across scanners and acquisition protocols. This finding underscores the need for segmentation methods that are not only accurate on benchmark datasets but also reproducible under the scanner heterogeneity characteristic of multi-site longitudinal studies, which motivated the external validation of HippoMamba on an independent multi-scanner cohort without retraining.

### 2.2 Longitudinal Neuroimaging and Atrophy Quantification

The pathophysiological cascade of AD unfolds over decades prior to clinical symptom onset, and the temporal dynamics of biomarker progression follow a well-characterized sequence in which hippocampal atrophy emerges as one of the earliest detectable structural changes [35]. Longitudinal MRI studies have established that the annual rate of hippocampal volume loss is higher in MCI and AD dementia than in cognitively normal older adults [36]. These atrophy rates, when measured reliably over serial MRI acquisitions, provide sensitive markers of disease progression and are increasingly used as outcome measures in clinical trials of disease-modifying therapies. Automated longitudinal analysis of hippocampal volumes has traditionally relied on cross-sectional application of segmentation tools such as FreeSurfer [6], with volume differences between timepoints computed as simple subtraction of independently derived estimates. This approach is suboptimal because it treats measurement noise at each timepoint as independent, whereas in reality the hippocampi of the same individual across serial visits share strong anatomical consistency that could be exploited to stabilize volume estimates. Longitudinal processing streams in FreeSurfer partially address this by constructing a within-subject template and propagating it across timepoints [37], but this approach remains computationally intensive.

The application of deep learning to longitudinal neuroimaging has primarily focused on disease progression prediction rather than joint segmentation and atrophy quantification. Recurrent architectures including LSTMs and gated recurrent units have been applied to model trajectories of cognitive assessments, CSF biomarkers, and imaging-derived phenotypes across serial visits [14,15]. Nguyen et al. [14] used deep recurrent networks trained on multimodal longitudinal ADNI data to forecast clinical diagnosis, cognition, and ventricular volume, while Ghazi et al. [15] showed that recurrent networks could be trained to handle missing data, a critical requirement for real-world longitudinal cohorts with irregular visit schedules. Predictive modeling of brain atrophy from longitudinal MRI has emerged as a distinct research direction with direct clinical relevance. Hadji et al. [16] showed that elastic net regression models using longitudinal MRI predicted future annual percentage change in hippocampal, ventricular, and total gray matter volume more accurately than models using a single timepoint, with cross-validated Pearson correlations of 0.62 for the hippocampus, 0.51 for the ventricles, and 0.41 for total gray matter when risk factors were included. Le et al. [17] proposed a U-Net-based model trained to predict two-year cortical atrophy rate maps from baseline MRI, demonstrating that deep learning can identify individuals with abnormal atrophy patterns from a single timepoint. Aghajanian et al. [18] further showed that combining longitudinal structural MRI features with radiomic descriptors improved prediction of AD progression in ADNI subjects, underscoring the complementary value of temporal and spatial information. Collectively, these studies motivate deep learning models with explicit temporal modeling capacity for serial MRI data, which informed the design of HippoMamba's Mamba-based temporal encoder. Standardized evaluation of MRI-based methods in dementia has been addressed by Bron et al. [38], who proposed a framework for benchmarking computer-aided diagnosis algorithms in dementia based on structural MRI, highlighting the importance of consistent preprocessing, harmonized cohort definitions, and appropriate statistical testing for fair comparison across methods. Dyrba et al. [39] combined grey matter volume with structural and functional connectivity measures in a multiple kernel classifier for AD, illustrating the role of volumetric measures within broader multimodal neuroimaging biomarker frameworks.

### 2.3 State Space Models in Medical Image Analysis

State space models (SSMs) represent a class of sequence modeling architectures grounded in control theory, in which the evolution of a latent state vector is governed by a linear dynamical system driven by an input sequence [40]. Classical SSMs such as the linear time-invariant system suffer from limited expressiveness and difficulty in capturing long-range dependencies, but the structured state space model S4 introduced by Gu et al. [40] addressed these limitations through a mathematically principled parameterization of the state transition matrix using the HiPPO framework, achieving competitive performance with transformers on long-range sequence benchmarks at a fraction of the computational cost. The Mamba architecture [19] extended S4 with a selective state space mechanism that conditions the SSM parameters on the input sequence, enabling dynamic and content-dependent state updates analogous to the gating mechanisms of LSTMs but retaining linear computational complexity. This selectivity allows Mamba to focus its representational capacity on the most informative elements of the input sequence, discarding irrelevant context while preserving long-range dependencies. In language modeling, Mamba demonstrated performance competitive with transformer-based models of comparable scale, with substantially higher inference throughput.

The translation of Mamba to visual and medical imaging tasks has proceeded rapidly. Vision Mamba [41] established the viability of pure SSM-based architectures for image classification and dense prediction by adapting a bidirectional scan strategy to two-dimensional image patches, while VM-UNet [22] built a U-shaped network from visual state space blocks for 2D medical image segmentation. U-Mamba [20] and SegMamba [21] proposed hybrid architectures combining convolutional feature extraction with Mamba-based global context modeling, reporting improved performance over convolutional and transformer baselines on tasks including abdominal organ segmentation [20] and 3D brain tumor segmentation [21]. More recently, SegMamba-V2 [25] extended this design to a broader range of 3D medical image segmentation datasets. Swin-UMamba† [42] showed that adapting an ImageNet-pretrained Mamba-based vision model through fine-tuning yields strong performance on medical image segmentation benchmarks, illustrating the benefit of large-scale pretraining for Mamba-based models. DCMamba [24] demonstrated the utility of Mamba for semi-supervised medical image segmentation through diversity-enhanced collaborative learning from unlabeled data. In computational pathology, MambaMIL [26] showed that sequence reordering strategies within Mamba encoders could improve multiple instance learning on gigapixel whole-slide images, highlighting the sensitivity of SSM-based models to the ordering of input elements, a property relevant to ordered inputs such as serial MRI visits. Notwithstanding these advances, the application of Mamba to longitudinal medical image analysis remains largely unexplored. Existing Mamba-based segmentation models operate on individual images or image patches, without modeling the temporal relationships between serial acquisitions from the same subject. HippoMamba applies a temporal Mamba encoder to the sequence of spatial feature representations extracted from serial MRI scans, so that segmentation and the prediction of hippocampal volume change between the first and last visits are trained within a single model.

## 3. Materials and Methods

### 3.1 Problem Formulation

Let $\mathcal{S} = \{I_1, I_2, \dots, I_T\}$ denote a longitudinal MRI sequence for a single subject, where $I_t \in \mathbb{R}^{D\times H\times W}$ represents the T1-weighted volumetric scan acquired at visit $t \in \{1, \dots, T\}$, with $D$, $H$, and $W$ denoting the spatial dimensions after preprocessing. For single-timepoint segmentation, a scan is associated with a hippocampal segmentation label $Y \in \{0,1,2\}^{D\times H\times W}$, where class 0 denotes background, class 1 denotes anterior hippocampus, and class 2 denotes posterior hippocampus, following the labeling convention of the Medical Segmentation Decathlon Task04 Hippocampus dataset [28]. Voxel-wise labels are not available for the longitudinal sequences. Instead, each longitudinal subject is associated with FreeSurfer-derived volumes $\mathbf{v} = (v_1^L, v_1^R, v_2^L, v_2^R, \dots, v_T^L, v_T^R) \in \mathbb{R}^{2T}$, where $v_t^L$ and $v_t^R$ denote the left and right hippocampal volumes in $mm^3$ at visit $t$, obtained from the cross-sectional FreeSurfer 6 processing distributed by ADNI as the UCSFFSX6 dataset [27]. Each sequence consists of the first visits, ordered by acquisition date, with a usable image and FreeSurfer volumes, up to a maximum of five, so that $T \leq 5$ and $v_T$ refers to the last of these visits. One subject had six usable visits. The longitudinal target for each hemisphere is defined as the fractional volume change between the first and last visits of the sequence:

$$r^L = \frac{v_T^L - v_1^L}{v_1^L + \epsilon}, \quad r^R = \frac{v_T^R - v_1^R}{v_1^R + \epsilon} \tag{1}$$

where $\epsilon = 10^{-8}$ is a small constant for numerical stability. This formulation produces a dimensionless fractional change in the range $(-1, \infty)$, with negative values indicating volume loss and positive values indicating apparent volume gain due to measurement variability. Because it is not divided by the interval between visits, it represents total change over the follow-up period rather than an annual rate. The vector $\mathbf{r} = (r^L, r^R) \in \mathbb{R}^2$ serves as the regression target for Phase 2 training. The learning objectives of HippoMamba are therefore twofold. In Phase 1, the model learns a mapping:

$$f_{\text{seg}}: \mathbb{R}^{D\times H\times W} \to \{0,1,2\}^{D\times H\times W} \tag{2}$$

that produces voxel-wise hippocampal segmentation from a single MRI volume. In Phase 2, the model learns an additional mapping:

$$f_{\text{traj}}: \mathbb{R}^{T\times D\times H\times W} \to \mathbb{R}^2 \tag{3}$$

that predicts the fractional volume change vector $\mathbf{r}$ from the longitudinal sequence $\mathcal{S}$, conditioned on the spatial representations learned during Phase 1.

### 3.2 HippoMamba Architecture

The HippoMamba architecture consists of four components, namely a 3D convolutional spatial encoder $\mathcal{E}_\phi$, a Mamba-based temporal encoder $\mathcal{T}_\psi$, a UNet-style segmentation decoder $\mathcal{D}_\xi$, and an atrophy head $\mathcal{A}_\omega$. For a sequence of $T$ visits, each volume $I_t$ is processed independently by the spatial encoder, which produces multi-scale feature maps and a pooled visit embedding. The temporal encoder processes the sequence of visit embeddings, and its output for each visit is injected into the deepest feature map of that visit before decoding into segmentation logits. A subject-level summary of the temporal encoder output is passed to the atrophy head to predict the fractional volume change. For a single scan, $T = 1$. The overall architecture is illustrated in Fig. 1.

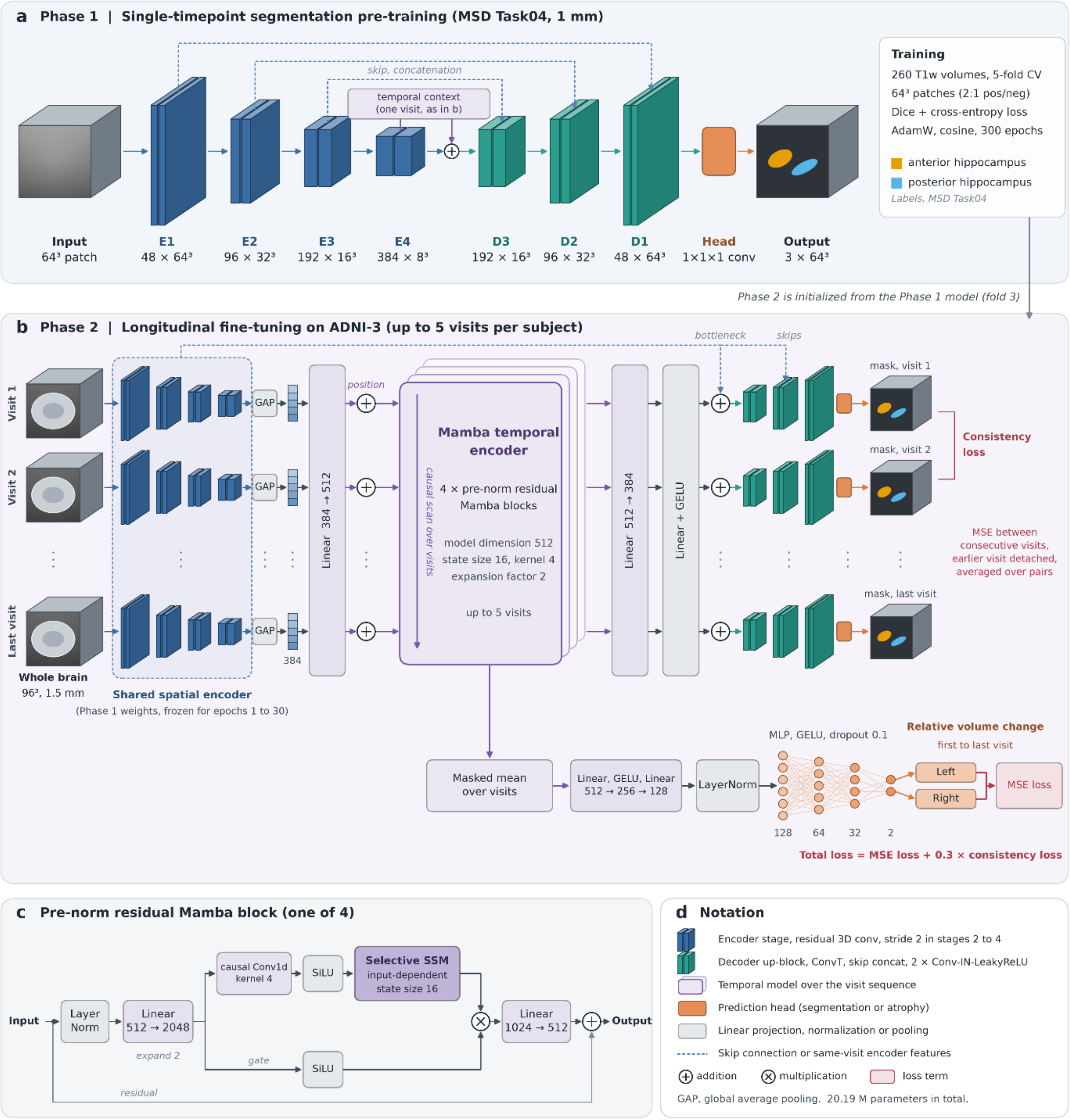

**Fig. 1.** HippoMamba architecture and two-phase training. (a) In Phase 1, the full network is trained with a single visit on 64³ patches of T1-weighted volumes from the Medical Segmentation Decathlon hippocampus task (1 mm) with a Dice and cross-entropy loss to segment the anterior and posterior hippocampus. A four-stage residual 3D convolutional encoder (48, 96, 192 and 384 channels, stride 2 in stages 2

to 4) feeds a three-level decoder through skip connections, and the temporal context computed from the single visit is added to the bottleneck features. (b) In Phase 2, the network is initialized from the Phase 1 model and fine-tuned on up to five whole-brain ADNI-3 visits per subject ($96^3$, 1.5 mm). The shared spatial encoder, frozen for the first 30 epochs, extracts features from each visit, which are globally average pooled, projected to 512 dimensions and summed with a learnable positional embedding. A temporal encoder of four pre-norm residual Mamba blocks processes the visit sequence. Each per-visit output is projected to 384 dimensions and added as a channel-wise bias to the bottleneck features of the same visit before decoding, and a consistency term, averaged over consecutive visit pairs, penalizes differences between their segmentation logits. The masked mean of the temporal outputs is projected to 128 dimensions, layer-normalized and passed to a three-layer multilayer perceptron that predicts the relative change in left and right hippocampal volume between the first and last visits. (c) Structure of one pre-norm residual Mamba block. (d) Notation. GAP, global average pooling.

### 3.2.1 Spatial Encoder

The spatial encoder $\mathcal{E}_\phi: \mathbb{R}^{1\times D\times H\times W} \to \mathbb{R}^{384\times\frac{D}{8}\times\frac{H}{8}\times\frac{W}{8}}$ is a four-stage 3D convolutional network with residual connections. Each stage $s \in \{1,2,3,4\}$ is a residual block of two 3×3×3 convolutions, each followed by instance normalization and LeakyReLU activation with negative slope 0.01, and a 1×1×1 convolution with instance normalization on the residual path whenever the number of channels or the resolution changes. The first convolution of stages 2 to 4 has stride 2, so that the spatial resolution is halved three times. The feature map at stage $s$ is computed as:

$$\mathrm{F}_t^{(s)} = \mathrm{ConvBlock}_s\left(\mathrm{F}_t^{(s-1)}\right) + \mathrm{Skip}_s\left(\mathrm{F}_t^{(s-1)}\right) \quad (4)$$

where $\mathrm{F}_t^{(0)} = I_t$ and the stages have feature dimensions $[C_1, C_2, C_3, C_4] = [48,96,192,384]$. The deepest feature map $\mathrm{z}_t = \mathrm{F}_t^{(4)} \in \mathbb{R}^{384\times\frac{D}{8}\times\frac{H}{8}\times\frac{W}{8}}$ is globally average-pooled to produce a visit embedding:

$$\mathbf{e}_t = \mathrm{AvgPool}(\mathbf{z}_t) \in \mathbb{R}^{384} \quad (5)$$

The encoder weights $\phi$ are shared across all visits, enforcing a consistent spatial representation regardless of visit index.

### 3.2.2 Mamba Temporal Encoder

The temporal encoder $\mathcal{T}_\psi$ receives the sequence of visit embeddings $\mathrm{E} = (\mathrm{e}_1, \mathrm{e}_2, \dots, \mathrm{e}_T) \in \mathbb{R}^{T\times 384}$, projects it to dimension $d_{\text{model}} = 512$, and adds learnable positional embeddings $\mathrm{P} \in \mathbb{R}^{T_{\max}\times d_{\text{model}}}$ to provide explicit information about visit ordering:

$$\tilde{\mathrm{E}} = \mathrm{E}\mathrm{W}_{\text{in}}^{\mathrm{T}} + \mathrm{b}_{\text{in}} + \mathrm{P}_{1:T} \quad (6)$$

where $\mathrm{W}_{\text{in}} \in \mathbb{R}^{d_{\text{model}}\times 384}$ and $T_{\max} = 5$ is the maximum number of visits supported by the architecture. The sequence is then processed by a stack of $L = 4$ Mamba blocks implementing the selective state space mechanism of Gu and Dao [19]. Within each block, the input is expanded to $d_{\text{inner}} = \beta d_{\text{model}}$ channels with expansion factor $\beta = 2$ and passed through a depthwise causal convolution of kernel size $d_{\text{conv}} = 4$, after which a selective state space model is applied independently to each channel. For a channel input $u_t$ at visit $t$, the hidden state and output are:

$$h_t = \overline{\mathrm{A}}_t h_{t-1} + \overline{\mathrm{B}}_t u_t \quad (7)$$

$$y_t = \mathrm{C}_t h_t + \mathrm{D} u_t \quad (8)$$

where $h_t \in \mathbb{R}^{d_{\text{state}}}$ with $d_{\text{state}} = 16$, and $\mathrm{B}_t$, $\mathrm{C}_t$ and the timescale $\Delta_t$ are obtained through learned projections of the input, which makes the state update input-dependent. The state matrix A is diagonal and initialized following the real-valued scheme of [19], which is derived from the HiPPO theory [40]. Discretization follows the reference implementation of [19]:

$$\overline{\mathrm{A}}_t = \exp(\Delta_t \mathrm{A}) \quad (9)$$

$$\overline{\mathrm{B}}_t = \Delta_t \mathrm{B}_t \quad (10)$$

The block output is gated, projected back to $d_{\text{model}}$ channels, and wrapped in a residual connection with layer normalization:

$$\mathrm{u}^{(l+1)} = \mathrm{u}^{(l)} + \mathrm{MambaBlock}^{(l)}\left(\mathrm{LayerNorm}\left(\mathrm{u}^{(l)}\right)\right) \quad (11)$$

The output of the temporal encoder is the sequence $\mathrm{H} = \left(h_1^{(L)}, h_2^{(L)}, \dots, h_T^{(L)}\right) \in \mathbb{R}^{T\times d_{\text{model}}}$. Each element is projected back to the encoder dimension to give a per-visit context vector $\mathrm{c}_t = \mathrm{W}_{\text{out}} h_t^{(L)} + \mathrm{b}_{\text{out}} \in \mathbb{R}^{384}$, and a subject-level representation is obtained by averaging over the valid visits:

$$\bar{\mathbf{h}} = \frac{1}{T}\sum_{t=1}^{T} h_t^{(L)} \in \mathbb{R}^{d_{\text{model}}} \tag{12}$$

The selective scan mechanism of the temporal encoder is illustrated in Fig. 2.

### 3.2.3 Segmentation Decoder

The segmentation decoder $\mathcal{D}_\xi$ follows the UNet decoder design with skip connections from the spatial encoder. For each visit $t$, the context vector $\mathrm{c}_t$ is transformed and added to the deepest encoder feature map as a channel-wise bias:

$$\hat{\mathrm{z}}_t = \mathrm{z}_t + \mathrm{GELU}(\mathrm{W}_c \mathrm{c}_t + \mathrm{b}_c)\mathrm{1}^{\mathsf{T}} \tag{13}$$

where $\mathrm{W}_c \in \mathbb{R}^{384\times384}$is a learned weight matrix and $\mathrm{1}^{\mathsf{T}}$ denotes broadcasting over all spatial positions. The fused feature map $\hat{\mathrm{z}}_t$ is decoded through three upsampling stages, each consisting of a transposed convolution with kernel size 2 and stride 2, concatenation with the corresponding encoder feature map, and two 3×3×3 convolutions with instance normalization and LeakyReLU activation:

$$\mathrm{G}_t^{(s-1)} = \mathrm{ConvBlock}_s\left(\mathrm{Cat}\left(\mathrm{Upsample}\left(\mathrm{G}_t^{(s)}\right), \mathrm{F}_t^{(s-1)}\right)\right), \quad s = 4,3,2 \tag{14}$$

where $\mathrm{G}_t^{(4)} = \hat{\mathrm{z}}_t$ and $\mathrm{Cat}(\cdot)$ denotes concatenation along the channel dimension. The final feature map $\mathrm{G}_t^{(1)} \in \mathbb{R}^{48\times D\times H\times W}$ is passed through a 1×1×1 convolution to produce per-class logits $\hat{Y}_t \in \mathbb{R}^{3\times D\times H\times W}$, which are converted to class probabilities through a softmax operation.

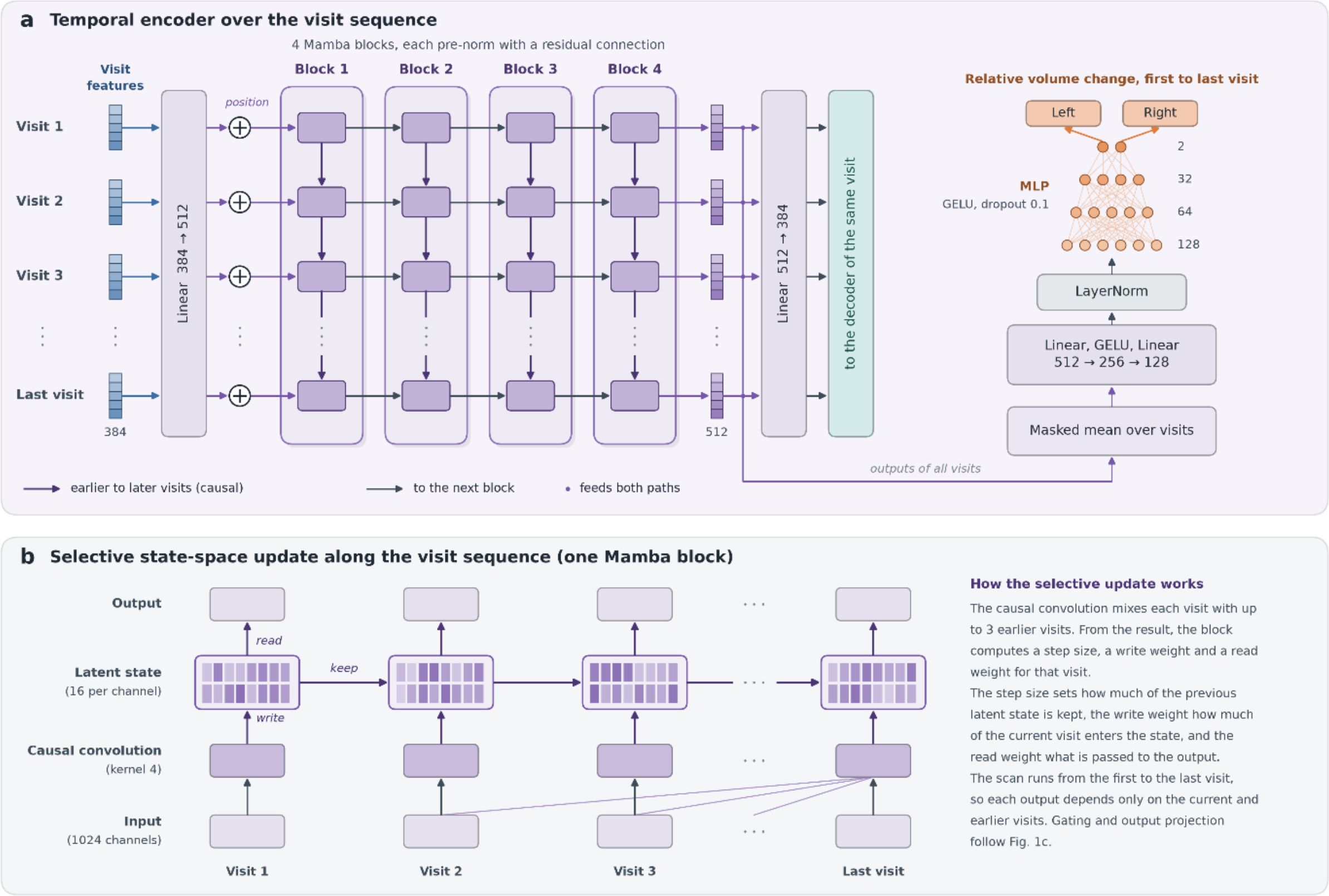

**Fig. 2.** Mamba temporal encoder. (a) For each visit, the globally average pooled bottleneck features of the spatial encoder (384 dimensions) are projected to 512 dimensions and summed with a learnable position embedding. Four pre-norm residual Mamba blocks process the visit sequence in temporal order, so that the output for each visit depends only on that visit and earlier visits. Each per-visit output is projected to 384 dimensions and passed to the decoder of the same visit. The masked mean of the outputs over all visits passes through two linear layers with a GELU activation (512, 256 and 128 dimensions), layer normalization and a three-layer multilayer perceptron that predicts the relative change in left and right hippocampal volume between the first and last visits. (b) Selective state-space update within one Mamba block. A causal convolution mixes each visit with up to three earlier visits, and an input-dependent step size, write weight and read weight then determine how much of the latent state is kept, how much of the current visit enters it and what is passed to the output.

### 3.2.4 Atrophy Head

The atrophy head $\mathcal{A}_\omega$ maps the subject-level representation $\bar{\mathrm{h}}$ to the predicted fractional volume change $\hat{\mathrm{r}} \in \mathbb{R}^2$. The representation is first reduced to an embedding $\mathrm{a} = \mathrm{W}_b\mathrm{GELU}(\mathrm{W}_a\bar{\mathrm{h}} + \mathrm{b}_a) + \mathrm{b}_b \in \mathbb{R}^{128}$, which is then passed through a multilayer perceptron:

$$\hat{\mathrm{r}} = \mathrm{W}_r\mathrm{GELU}(\mathrm{W}_2\mathrm{Dropout}(\mathrm{GELU}(\mathrm{W}_1\mathrm{LayerNorm}(\mathrm{a}) + \mathrm{b}_1)) + \mathrm{b}_2) + \mathrm{b}_r \tag{15}$$

where $\mathrm{W}_a \in \mathbb{R}^{256\times512}$, $\mathrm{W}_b \in \mathbb{R}^{128\times256}$, $\mathrm{W}_1 \in \mathbb{R}^{64\times128}$, $\mathrm{W}_2 \in \mathbb{R}^{32\times64}$ and $\mathrm{W}_r \in \mathbb{R}^{2\times32}$ are learned weight matrices, and dropout with probability $p = 0.1$ is applied once. The output $\hat{\mathrm{r}} = (\hat{r}^L, \hat{r}^R)$represents the predicted fractional volume changes for the left and right hippocampi. The head also contains a scalar output intended for conversion risk, which was not trained or used in this study.

### 3.3 Loss Functions

#### 3.3.1 Phase 1: Segmentation Loss

During Phase 1 training, each MRI volume is processed independently as a single-timepoint input with $T = 1$, so that the temporal encoder operates on a sequence of length one. The segmentation loss is a combination of Dice loss and cross-entropy loss with equal weights:

$$\mathcal{L}_{\mathrm{seg}} = \lambda_{\mathrm{Dice}}\mathcal{L}_{\mathrm{Dice}} + \lambda_{\mathrm{CE}}\mathcal{L}_{\mathrm{CE}} \tag{16}$$

where $\lambda_{\mathrm{Dice}} = \lambda_{\mathrm{CE}} = 1.0$. The Dice loss is computed for each class and averaged over the $K = 3$ classes, including background [43]:

$$\mathcal{L}_{\mathrm{Dice}} = 1 - \frac{1}{K}\sum_{k=0}^{K-1}\frac{2\sum_i \hat{p}_{i,k}\, y_{i,k} + \epsilon}{\sum_i \hat{p}_{i,k} + \sum_i y_{i,k} + \epsilon} \tag{17}$$

where $\hat{p}_{i,k}$ is the predicted softmax probability for voxel $i$ and class $k$, $y_{i,k}$ is the corresponding one-hot encoded label, and $\epsilon = 10^{-5}$ is a smoothing constant. The cross-entropy loss is the standard categorical cross-entropy over all voxels and classes:

$$\mathcal{L}_{\mathrm{CE}} = -\frac{1}{N}\sum_{i=1}^{N}\sum_{k=0}^{K-1} y_{i,k}\log\hat{p}_{i,k} \tag{18}$$

where $N$ is the total number of voxels in the input volume. Both terms were computed with the MONAI implementation.

#### 3.3.2 Phase 2: Longitudinal Loss

During Phase 2 training, the longitudinal sequence of $T \geq 2$ visits is processed jointly, and the model is optimized to predict the FreeSurfer-derived fractional volume changes. The volume change loss is a mean squared error between the predicted and reference fractional volume changes:

$$\mathcal{L}_{\mathrm{traj}} = \frac{1}{2}\sum_{h\in\{L,R\}}(\hat{r}^h - r^h)^2 \tag{19}$$

where $\hat{r}^h$ and $r^h$ denote the predicted and reference fractional volume changes for hemisphere $h \in \{L, R\}$. A longitudinal consistency term encourages similar segmentation outputs across consecutive visits:

$$\mathcal{L}_{\mathrm{cons}} = \frac{1}{T-1}\sum_{t=2}^{T}\frac{1}{N}\left\|\hat{Y}_t - \mathrm{sg}(\hat{Y}_{t-1})\right\|_2^2 \tag{20}$$

where $\hat{Y}_t$ denotes the segmentation logits at visit $t$, $N$ is the number of logit elements, and $\mathrm{sg}(\cdot)$ denotes the stop-gradient operator. The total Phase 2 loss is:

$$\mathcal{L}_{\mathrm{Phase\ 2}} = \mathcal{L}_{\mathrm{traj}} + \lambda_{\mathrm{cons}}\mathcal{L}_{\mathrm{cons}}, \qquad \lambda_{\mathrm{cons}} = 0.3 \tag{21}$$

The mean squared error formulation was chosen because the FreeSurfer volumes provide continuous, interval-scale measurements, and it directly penalizes deviations in the magnitude of the predicted change.

### 3.4 Two-Phase Training Strategy

### 3.4.1 Phase 1: Single-Timepoint Segmentation

In Phase 1, HippoMamba is trained as a single-timepoint segmentation model on the Medical Segmentation Decathlon Task04 Hippocampus dataset [28], using a five-fold cross-validation protocol with the same fold splits as the comparison baselines. Each fold consists of 208 training subjects and 52 validation subjects. All components of the model, including the temporal encoder, are trained jointly from random initialization using the AdamW optimizer [44] with an initial learning rate of $\eta_0 = 10^{-4}$, weight decay $\lambda = 10^{-5}$, and a batch size of one volume. The learning rate follows a cosine annealing schedule over $E_1 = 300$ epochs with a minimum learning rate of $\eta_{\min} = 10^{-6}$. Training uses mixed precision, and gradient norm clipping with a threshold of 1.0 is applied to the gradients scaled by the mixed-precision loss scaler before each optimizer step. Each volume is resampled to 1 mm isotropic spacing, reoriented to RAS, clipped below zero, scaled to the 0.5th to 99.5th intensity percentiles, and normalized to zero mean and unit variance over non-zero voxels. Training patches of 64×64×64 voxels are sampled with a positive to negative ratio of 2 to 1, two patches per volume. Data augmentation includes random flipping along each spatial axis and random 90-degree rotation (probability 0.5 each), random rotation within ±30 degrees and random zoom in the range 0.85 to 1.15 (probability 0.3 each), Gaussian smoothing (probability 0.2), random intensity scaling by up to 30% and shifting by up to 0.1 (probability 0.3 each), Gaussian noise with standard deviation 0.1 (probability 0.2), and random gamma adjustment in the range 0.7 to 1.5 (probability 0.3). Every five epochs, the model is evaluated on the validation subjects, each padded to 64×64×64 voxels as a whole image, using the Dice coefficient averaged over the anterior and posterior labels, and the checkpoint with the highest validation Dice is retained. The random seed is 42. In an additional evaluation reported separately, test-time augmentation is applied using eight mirroring transformations corresponding to all combinations of flips along the three spatial axes, and the resulting softmax probability maps are averaged before argmax decoding. Although the full HippoMamba architecture, including the Mamba temporal encoder, remains active during Phase 1, each training sample consists of a single timepoint ( $T = 1$ ), so the temporal encoder processes a sequence of length one, contributing no cross-visit context but receiving gradient updates through the segmentation loss that provide an initialization for Phase 2.

### 3.4.2 Phase 2: Longitudinal Fine-Tuning

In Phase 2, the full HippoMamba model is initialized from the Phase 1 model of fold 3, which had the highest validation Dice of the five Phase 1 folds. The inputs are the complete skull-stripped and registered ADNI-3 volumes of 96×96×96 voxels at 1.5 mm resolution, normalized to zero mean and unit variance over non-zero voxels, and no data augmentation is applied. The spatial encoder $\mathcal{E}_\phi$ is frozen for the first $E_f = 30$ epochs, so that the temporal encoder $\mathcal{T}_\psi$, the decoder $\mathcal{D}_\xi$ and the atrophy head $\mathcal{A}_\omega$ adapt to the longitudinal data without changing the spatial features learned in Phase 1. During this period, the trainable parameters are optimized with AdamW at a learning rate of $10^{-5}$ and weight decay $10^{-5}$ under a cosine schedule. At epoch $E_f$, the spatial encoder is unfrozen, and all parameters are optimized with a new AdamW optimizer at the same learning rate and a cosine schedule over the remaining 120 epochs. Training proceeds for $E_2 = 150$ epochs, with subjects randomly shuffled at each epoch and a batch size of one subject. For each subject, up to $T_{\max} = 5$ visits are used in chronological order, so that sequences have their own length and no padding is required. Training uses mixed precision with gradient norm clipping at a threshold of 0.5 applied to the loss-scaled gradients of the trainable parameters, and the random seed is 42. Every 25 epochs, the model is evaluated on held-out validation subjects using the mean of the left and right Pearson correlations between predicted and reference fractional volume change, and the checkpoint with the highest value is retained.

### 3.5 Algorithm 1: HippoMamba Training Procedure

Algorithm 1: HippoMamba Two-Phase Training

Require: Decathlon dataset $\mathcal{D}_{\text{dec}}$, ADNI-3 dataset $\mathcal{D}_{\text{adni}}$, folds $K = 5$,
epochs $E_1 = 300$, $E_2 = 150$, freeze epochs $E_f = 30$, learning rates $\eta_0 = 10^{-4}$ (Phase 1) and $\eta_2 = 10^{-5}$ (Phase 2),
consistency weight $\lambda_{\text{cons}} = 0.3$
Ensure: Trained model parameters $\{\phi^*, \psi^*, \xi^*, \omega^*\}$
Phase 1: Single-Timepoint Segmentation
1: for $k = 0$ to $K - 1$ do
2: Initialize $\phi, \psi, \xi, \omega$ randomly
3: for $e = 1$ to $E_1$ do
4: for each patch $(I, Y)$ sampled from $\mathcal{D}_{\text{dec}}^{\text{train},k}$ **do**
5: $\{\mathbf{F}^{(s)}\}, \mathbf{e} \leftarrow \mathcal{E}_\phi(I)$ ▷ Eq. (4), (5)
6: $\mathbf{c}, \bar{\mathbf{h}} \leftarrow \mathcal{T}_\psi(\mathbf{e})$ with $T = 1$ ▷ Eq. (6)–(12)
7: $\hat{Y} \leftarrow \mathcal{D}_\xi(\{\mathbf{F}^{(s)}\}, \mathbf{c})$ ▷ Eq. (13), (14)
8: $\mathcal{L} \leftarrow \mathcal{L}_{\text{seg}}(\hat{Y}, Y)$ ▷ Eq. (16)–(18)
9: Update all parameters via AdamW with cosine annealing from $\eta_0$
10: end for

if $e \mod 5 = 0$ then evaluate validation Dice and keep the best checkpoint
end for
end for

Phase 2: Longitudinal Fine-Tuning

Initialize $\phi, \psi, \xi, \omega$ from the Phase 1 model of fold 3
Freeze $\phi$, and optimize $\psi, \xi, \omega$ via AdamW at $\eta_2$ with cosine annealing
for $e = 1$ to $E_2$ do
if $e = E_f + 1$ then unfreeze $\phi$, and optimize all parameters via a new AdamW at $\eta_2$ with cosine annealing over $E_2 - E_f$ epochs
for each sequence $\mathcal{S} = (I_1, \ldots, I_T) \in \mathcal{D}_{\text{adni}}^{\text{train}}$ in random order **do**
$\left\{\mathbf{F}_t^{(s)}\right\}, \mathbf{e}_t \leftarrow \mathcal{E}_\phi(I_t)$ for $t = 1, \ldots, T$ $\triangleright$ Eq. (4), (5)
$\{\mathbf{c}_t\}, \bar{\mathbf{h}} \leftarrow \mathcal{T}_\psi(\mathbf{e}_1, \ldots, \mathbf{e}_T)$ $\triangleright$ Eq. (6)–(12)
$\hat{Y}_t \leftarrow \mathcal{D}_\xi\left(\left\{\mathbf{F}_t^{(s)}\right\}, \mathbf{c}_t\right)$ for $t = 1, \ldots, T$ $\triangleright$ Eq. (13), (14)
$\hat{\mathbf{r}} \leftarrow \mathcal{A}_\omega(\bar{\mathbf{h}})$ $\triangleright$ Eq. (15)
$\mathcal{L} \leftarrow \mathcal{L}_{\text{traj}}(\hat{\mathbf{r}}, \mathbf{r}) + \lambda_{\text{cons}}\mathcal{L}_{\text{cons}}(\hat{Y}_1, \ldots, \hat{Y}_T)$ $\triangleright$ Eq. (19)–(21)
Update the trainable parameters via AdamW
end for
if $e \mod 25 = 0$ then evaluate the validation Pearson correlation and keep the best checkpoint
end for
return $\{\phi^*, \psi^*, \xi^*, \omega^*\}$ with the highest validation Pearson correlation

## 4. Experimental Setup

### 4.1 Datasets

#### 4.1.1 Medical Segmentation Decathlon Task04 Hippocampus

Phase 1 training and evaluation are conducted on the publicly available Medical Segmentation Decathlon (MSD) Task04 Hippocampus dataset [28], which constitutes a well-established benchmark for volumetric hippocampal segmentation in the medical image analysis community. The dataset comprises 260 labeled T1-weighted MRI volumes acquired at a single center from healthy adults and adults with a non-affective psychotic disorder, each manually annotated with voxel-level labels for the anterior hippocampus (hippocampal head) and the posterior hippocampus (hippocampal body and tail), yielding a three-class labeling scheme consisting of background, anterior hippocampus, and posterior hippocampus. An additional 130 test cases are provided with withheld ground truth labels as part of the original MSD challenge, and these cases are not used in any experiment reported in this paper. Each volume is a cropped region around the hippocampus at 1 mm isotropic resolution, with dimensions of up to 43, 59, and 47 voxels along the three spatial axes. The dataset does not include longitudinal acquisitions and is used exclusively for Phase 1 single-timepoint segmentation training and evaluation. A five-fold cross-validation protocol is applied to the 260 labeled cases, partitioning them into five non-overlapping folds of 52 subjects each, yielding 208 training and 52 validation subjects per fold. The fold assignments are identical for HippoMamba and all compared baseline methods, such that every method is trained on the same 208 subjects and evaluated on the same 52 subjects in each of the five cross-validation iterations. This protocol ensures that performance differences between methods are not caused by differences in evaluation subsets. For every method, the checkpoint with the highest validation Dice across training epochs is used for evaluation, consistent with standard model selection practice. Because this selection uses the same validation subjects on which performance is reported, the cross-validation results of all methods are expected to be slightly optimistic.

#### 4.1.2 ADNI-3 Longitudinal Dataset

Phase 2 longitudinal fine-tuning and evaluation are conducted on a curated subset of the Alzheimer's Disease Neuroimaging Initiative Phase 3 (ADNI-3) dataset [7], a large-scale multi-site longitudinal neuroimaging study designed to develop validated biomarkers for the early detection and tracking of AD. From the full ADNI-3 cohort, 271 subjects with at least one available preprocessed T1-weighted MRI scan and corresponding FreeSurfer hippocampal volumes passing the hippocampal quality control criterion (HippoQC = Pass) are identified, yielding 566 subject-visit pairs, all of which pass the quality control criterion. Among the 271 subjects, 182 have two or more MRI acquisitions and form the longitudinal cohort used for Phase 2 training and evaluation, and the remaining 89 single-visit subjects are excluded because a second timepoint is required to compute volume change. The longitudinal cohort comprises 82 cognitively normal (CN) subjects, 71 subjects with mild cognitive impairment (MCI), and 29 subjects with AD dementia at the baseline visit. Of these subjects, 96 have two visits, 62 three, 22 four, and 2 five visits used in the analysis, with a mean of 2.6 visits per subject and a follow-up between the first and last visit of 321 to 1,898 days (mean 2.60 years). Phase 2 is evaluated with subject-level five-fold cross-validation stratified by baseline diagnosis, in which 20% of each training portion is held out for checkpoint selection, so that every subject is predicted exactly once by a model that has not seen them during training or checkpoint selection. The demographic and clinical characteristics of the longitudinal cohort are summarised in Table 1. ADNI data were obtained from the Laboratory of Neuro Imaging Image and Data Archive (https://ida.loni.usc.edu) under an approved data use agreement with the ADNI Data Repository. The ADNI was launched in 2003 as a public-private partnership led by Principal Investigator Michael W. Weiner, MD, with the primary goal of testing whether serial MRI, positron emission tomography, other biological markers, and clinical and neuropsychological assessments can be combined to measure the progression of MCI and early AD [7].

**Table 1.** Demographic and clinical characteristics of the ADNI-3 longitudinal cohort (N = 182)

| Characteristic | CN (N = 82) | MCI (N = 71) | AD (N = 29) | Total (N = 182) |
|---|---|---|---|---|
| Age, years | 72.1 ± 6.1 | 73.1 ± 7.4 | 76.8 ± 8.2 | 73.2 ± 7.1 |
| Female, n (%) | 52 (63.4) | 34 (47.9) | 13 (44.8) | 99 (54.4) |
| MMSE at baseline | 29.2 ± 1.0 | 27.7 ± 2.0 | 22.8 ± 3.3 | 27.6 ± 2.9 |
| Visits per subject | 2.6 ± 0.7 (2 to 4) | 2.8 ± 0.8 (2 to 5) | 2.3 ± 0.5 (2 to 3) | 2.6 ± 0.7 (2 to 5) |
| Follow-up, years | 3.09 ± 1.13 (1.04 to 5.20) | 2.43 ± 1.20 (0.88 to 4.64) | 1.62 ± 0.71 (0.96 to 3.37) | 2.60 ± 1.22 (0.88 to 5.20) |
| FreeSurfer baseline hippocampal volume, mm³ | 7699 ± 797 | 7069 ± 1099 | 5656 ± 1078 | 7128 ± 1195 |

*CN, cognitively normal. MCI, mild cognitive impairment. AD, Alzheimer's disease dementia. MMSE, Mini-Mental State Examination. Values are mean ± standard deviation with ranges in parentheses, unless otherwise stated. Diagnosis, age and MMSE are taken at the baseline visit. Visits per subject counts the visits used, up to five. Follow-up is the interval between the first and last visit used. Hippocampal volume is the sum of the left and right FreeSurfer volumes at baseline.*

#### 4.1.3 UCSFFSX6 FreeSurfer Reference Volumes

Hippocampal volumes serving as the reference for Phase 2 training and evaluation are taken from the UCSFFSX6 dataset distributed by ADNI, which contains the output of cross-sectional FreeSurfer 6 processing [6], [27] of the ADNI-3 T1-weighted MRI scans, in which each session is segmented independently. Left and right hippocampal volumes in mm³ are extracted for each subject-visit combination that passes the hippocampal quality control criterion, yielding 566 subject-visit pairs with valid volumetric measurements across the full 271-subject cohort. These measurements form the ADNI longitudinal manifest used for Phase 2 training supervision and evaluation. All investigators within the ADNI contributed to the design and implementation of the study and provided data but did not participate in analysis or writing of this report.

### 4.2 Data Preprocessing

#### 4.2.1 Phase 1 Preprocessing

All MSD Task04 Hippocampus volumes undergo a standardized preprocessing pipeline prior to model training and evaluation, implemented with the MONAI medical image analysis framework. Each volume is first resampled to isotropic 1.0 mm voxel spacing, using linear interpolation for images and nearest-neighbor interpolation for labels, and reoriented to the RAS (right-anterior-superior) coordinate convention. Intensity normalization is applied through a three-step procedure. Voxel intensities below zero are first set to zero, intensities are then linearly rescaled to the range [0, 1] between the 0.5th and 99.5th percentiles of the volume, with values outside this range clipped, and a final z-score normalization is applied over non-zero voxels to produce zero-mean, unit-variance intensities. During training, volumes are padded to a minimum extent of 64×64×64 voxels, and random crops of 64×64×64 voxels are sampled with a positive to negative ratio of 2 to 1, so that two thirds of the sampled patches are centred on hippocampal voxels, which mitigates class imbalance between hippocampal and background voxels. Two patches are sampled per volume. Data augmentation is applied online during training using stochastic transformations, each applied independently with a specified probability. These comprise random flipping along each of the three spatial axes with probability 0.5, random 90-degree rotation with probability 0.5, random rotation within ±30 degrees around each axis with probability 0.3, random zoom in the range 0.85 to 1.15 with probability 0.3, Gaussian smoothing with standard deviation sampled from 0.5 to 1.0 voxels with probability 0.2, random intensity scaling by a factor sampled from 0.7 to 1.3 with probability 0.3, random intensity shifting by an offset sampled from −0.1 to 0.1 with probability 0.3, additive Gaussian noise with standard deviation 0.1 with probability 0.2, and random gamma contrast adjustment with gamma sampled from 0.7 to 1.5 with probability 0.3. This augmentation strategy is intended to approximate the variability of acquisition characteristics present in multi-site neuroimaging datasets. For validation, each volume is padded to 64×64×64 voxels and evaluated as a whole image. In the additional evaluation reported separately, test-time augmentation is applied to HippoMamba using all eight mirroring transformations corresponding to the complete set of axis-aligned reflections, and the resulting softmax probability maps are averaged elementwise before argmax decoding to the final segmentation label map.

#### 4.2.2 Phase 2 Preprocessing

The ADNI-3 T1-weighted MRI volumes were preprocessed before this study by brain extraction using Otsu thresholding with morphological cleanup, affine registration to the MNI152 1 mm template using ANTsPyX with SimpleITK as a fallback, z-score intensity normalization within the brain mask, and resampling and centre-cropping to 96×96×96 voxels at 1.5 mm isotropic resolution, yielding whole-brain volumes on an identical grid for all subjects. For HippoMamba, each volume is normalized again to zero mean and unit variance over non-zero voxels, and the complete volume is used as input without cropping, so that both hippocampi are retained in full. Visits are ordered chronologically by the visit number recorded in the ADNI longitudinal manifest, and visit indices are assigned sequentially from 1 to $T$ regardless of the actual inter-visit interval in days. No temporal interpolation or visit imputation is performed. Because each training step processes the sequence of a single subject, sequences are not padded, and the subject-level representation in Equation (12) is the mean over all visits of the subject.

### 4.3 Comparison Methods

#### 4.3.1 Single-Timepoint Baselines

HippoMamba is compared against six single-timepoint segmentation methods spanning CNN-based and transformer-based architectures, all trained and evaluated under the same five-fold cross-validation protocol and MSD Task04 Hippocampus fold assignments, with a learning rate of $10^{-4}$, weight decay $10^{-5}$, and early stopping with a patience of 25 and a maximum of 200 epochs. All baseline models are trained with a spatial patch size of 64×64×64 and a batch size of 2, without test-time augmentation. The 3D U-Net [10] is the MONAI U-Net with five resolution levels, feature dimensions of 32, 64, 128, 256, and 320, two residual units per level, and dropout of 0.1. UNETR [11] uses a Vision Transformer encoder operating on non-overlapping 3D patches of size $16^3$with hidden dimension 768 and a convolutional decoder with feature size 16. SegResNet [32] is a fully convolutional residual encoder-decoder with an initial filter count of 8 and progressive channel doubling. MedNeXt [33] implements the ConvNeXt design for 3D medical image segmentation, and the small (S) variant is used with kernel size 3 and expansion ratio 2. DynUNet [34] is a dynamic U-Net implementation in MONAI with kernel sizes of $3^3$ at all five levels, strides of 1, 2, 2, 2, and 2, filter progression of 32, 64, 128, 256, and 320, and no deep supervision. Attention U-Net [31] extends the 3D U-Net with additive attention gates at the decoder skip connections, using feature dimensions of 32, 64, 128, and 256 and strides of 2, 2, and 2. UNETR, SegResNet, MedNeXt, and Attention U-Net were trained in a previous study with the Adam optimizer [45] at a constant learning rate, z-score normalization, and padding without resampling or percentile rescaling, whereas the 3D U-Net and DynUNet were trained with the AdamW optimizer [44], cosine annealing, and the same preprocessing as HippoMamba (Section 4.2.1). HippoMamba differs from all baselines in being trained for 300 epochs without early stopping and with a batch size of one volume.

### 4.4 Evaluation Metrics

#### 4.4.1 Segmentation Metrics

Segmentation performance is assessed using three complementary metrics that capture different aspects of volumetric overlap and boundary accuracy, consistent with the standard evaluation protocol for hippocampal segmentation benchmarks [13], [28]. The Dice Similarity Coefficient (DSC) [43] is defined as:

$$\mathrm{DSC} = \frac{2|P \cap G|}{|P| + |G|} \tag{22}$$

where $P$ and $G$ denote the predicted and reference segmentation masks respectively, and it measures volumetric overlap and serves as the primary ranking metric. For the surface metrics, let $S_P$ and $S_G$ denote the surface voxels of the predicted and reference segmentations, and let $d(x, S) = \min_{s \in S} \|x - s\|_2$ denote the distance from a point $x$ to a surface $S$. The 95th percentile Hausdorff Distance (HD95) [46] is defined as:

$$\mathrm{HD95} = \max\left(\mathrm{perc}_{95}\{d(p, S_G)\}_{p \in S_P},\ \mathrm{perc}_{95}\{d(g, S_P)\}_{g \in S_G}\right) \tag{23}$$

and measures the near-worst-case boundary deviation in millimeters while discarding the most extreme 5 percent of surface distances. The Average Surface Distance (ASD) measures the mean symmetric distance between the predicted and reference surfaces:

$$\mathrm{ASD} = \frac{1}{|S_P| + |S_G|}\left(\sum_{p \in S_P} d\,(p, S_G) + \sum_{g \in S_G} d\,(g, S_P)\right) \tag{24}$$

All three metrics are computed for each subject separately for the anterior and posterior hippocampal subregions with the MONAI implementation and averaged across both subregions, then averaged within each fold. Five-fold cross-validation results are reported as mean ± standard deviation across the five-fold means.

#### 4.4.2 Volumetric Agreement Metrics

Agreement between hippocampal volumes or volume changes derived from HippoMamba and the FreeSurfer reference is assessed using the Pearson and Spearman correlation coefficients, the intraclass correlation coefficients ICC(3,1) for consistency and ICC(2,1) for absolute agreement [47], [48], the mean absolute error and mean absolute percentage error, and Bland-Altman analysis [49], with bias and limits of agreement computed as:

$$\text{Bias} = \frac{1}{N}\sum_{i=1}^{N}(\hat{x}_i - x_i) \tag{25}$$

$$\text{LoA} = \text{Bias} \pm 1.96 \times \text{SD}(\hat{x}_i - x_i) \tag{26}$$

where $\hat{x}_i$ and $x_i$ denote the HippoMamba-derived and reference values for subject or session $i$, $N$ is the number of subjects or sessions evaluated, and SD denotes the standard deviation of the pairwise differences. ICC values are interpreted following the guidelines of Koo and Li [47], with values below 0.50 indicating poor, 0.50 to 0.75 moderate, 0.75 to 0.90 good, and above 0.90 excellent reliability. Confidence intervals are obtained from a participant-level bootstrap with 2,000 resamples. For the longitudinal module, predictions of fractional volume change are additionally compared with those of a linear model based only on follow-up duration and number of visits, fitted within the same cross-validation folds, and with the partial correlation between predicted and reference change after adjustment for these two variables.

### 4.5 Implementation Details

All experiments are implemented in Python using PyTorch and the MONAI medical image analysis framework, and all models are trained on a single NVIDIA GeForce RTX 4060 Laptop GPU with 8 GB VRAM. HippoMamba is trained with Python 3.11.17, PyTorch 2.5.1, CUDA 12.1, MONAI 1.3.0, and the mamba-ssm library 2.3.2, which provides hardware-optimized CUDA kernels for the selective state space computation. Evaluation of all methods and all statistical analyses are performed with Python 3.12.13, PyTorch 2.11.0, CUDA 12.8, MONAI 1.6.0, mamba-ssm 2.3.2, and SciPy 1.18.1. The 3D U-Net, UNETR, SegResNet, DynUNet, and Attention U-Net use their MONAI implementations, and MedNeXt uses the implementation of its authors. HippoMamba is trained with mixed precision using PyTorch automatic mixed precision with dynamic loss scaling, and gradient-norm clipping with a threshold of 1.0 in Phase 1 and 0.5 in Phase 2 is applied to the loss-scaled gradients before each optimizer step. The 3D U-Net and DynUNet are also trained with mixed precision, whereas UNETR, SegResNet, MedNeXt, and Attention U-Net are trained in full precision. HippoMamba contains 20.19 million parameters, with spatial encoder feature dimensions of 48, 96, 192, and 384, Mamba hidden dimension 512, state dimension 16, convolutional kernel size 4, expansion factor 2, and 4 Mamba layers. Phase 1 training requires 3.1 to 3.7 hours per fold (16.7 hours for all five folds), and Phase 2 training requires approximately 3 to 5.5 hours per cross-validation fold. Code and model weights will be made publicly available upon acceptance.

### 4.6 Statistical Analysis

All statistical analyses are performed with SciPy 1.18.1 and NumPy, using analysis scripts fixed before the corresponding results were examined. For the segmentation comparison, per-subject Dice, HD95, and ASD values of each baseline are compared with those of HippoMamba without test-time augmentation using two-sided Wilcoxon signed-rank tests on the 260 paired subjects, and the resulting p-values are adjusted with the Holm procedure across the six baselines for each metric, with significance at $\alpha = 0.05$. Results with test-time augmentation are reported descriptively without significance testing. Intraclass correlation coefficients are computed from two-way analysis of variance mean squares following Shrout and Fleiss [48], with ICC (3,1) for consistency and ICC (2,1) for absolute agreement. Confidence intervals for correlation coefficients, intraclass correlation coefficients, and Bland-Altman bias are obtained from a participant-level bootstrap with 2,000 resamples and a fixed random seed, so that all sessions or hemispheres of a participant are resampled together. For the OASIS-3 external validation, sessions are included according to a registration quality criterion fixed before evaluation of the reported models, and the analysis is repeated with two alternative thresholds as a sensitivity analysis. For the longitudinal module, the primary measure is the Pearson correlation between predicted and reference fractional volume change over all cross-validated subjects. It is compared with a linear model based only on follow-up duration and number of visits fitted within the same folds, with a two-sided bootstrap p-value for the difference in correlation, and is complemented by the partial correlation after adjustment for these two variables. All reported p-values are two-tailed.

## 5. Results

### 5.1 Phase 1: Hippocampal Segmentation Performance

### 5.1.1 Quantitative Comparison

Table 2 compares HippoMamba with six baselines on the Medical Segmentation Decathlon Task04 hippocampus dataset under five-fold cross-validation, with all methods evaluated on the same 260 validation subjects by a single evaluation procedure. Without test-time augmentation, HippoMamba obtained the highest mean Dice among the compared methods (0.8861 ± 0.0094), followed by Attention U-Net (0.8835 ± 0.0087) and SegResNet (0.8737 ± 0.0083). Paired per-subject tests showed significantly higher Dice for HippoMamba than for every baseline ($p < 0.0001$ against UNETR, 3D U-Net, DynUNet, MedNeXt and SegResNet, and $p = 0.0130$ against Attention U-Net). The margin over Attention U-Net was small, with a median paired difference of 0.0015, and HippoMamba was higher in four of the five folds (Table 3). HippoMamba also had the lowest mean HD95 (1.4533 ± 0.1022 mm) and ASD (0.4430 ± 0.0340 mm). These boundary metrics did not differ significantly from those of Attention U-Net ($p = 0.8107$ for HD95 and $p = 0.0737$ for ASD), and HD95 did not differ significantly from that of SegResNet ($p = 0.1735$), whereas the ASD of SegResNet and both boundary metrics of the remaining four baselines were significantly higher ($p < 0.0001$). With eight-flip test-time augmentation, the mean Dice of HippoMamba increased to 0.8888 ± 0.0095, and HD95 and ASD decreased to 1.4333 ± 0.0925 mm and 0.4332 ± 0.0357 mm. UNETR [11], the only transformer-based baseline, had the lowest Dice (0.8597 ± 0.0088) and the highest HD95 (1.8675 ± 0.1314 mm) of all methods.

**Table 2.** Cross-validation performance on the Medical Segmentation Decathlon hippocampus dataset (Task04, n = 260). Metrics are mean ± standard deviation across five folds. Wilcoxon signed-rank test p-values compare each baseline to HippoMamba without test-time augmentation on per-subject scores, with Holm adjustment across the six baselines for each metric.

| Method | Dice ↑ | HD95 (mm) ↓ | ASD (mm) ↓ | p (Dice) | p (HD95) | p (ASD) |
|---|---|---|---|---|---|---|
| UNETR | 0.8597 ± 0.0088 | 1.8675 ± 0.1314 | 0.5507 ± 0.0361 | < 0.0001 | < 0.0001 | < 0.0001 |
| 3D U-Net | 0.8605 ± 0.0085 | 1.6242 ± 0.1333 | 0.5528 ± 0.0303 | < 0.0001 | < 0.0001 | < 0.0001 |
| DynUNet | 0.8639 ± 0.0082 | 1.7072 ± 0.1161 | 0.5364 ± 0.0268 | < 0.0001 | < 0.0001 | < 0.0001 |
| MedNeXt | 0.8641 ± 0.0101 | 1.7291 ± 0.2139 | 0.5298 ± 0.0372 | < 0.0001 | < 0.0001 | < 0.0001 |
| SegResNet | 0.8737 ± 0.0083 | 1.5127 ± 0.0994 | 0.4892 ± 0.0264 | < 0.0001 | 0.1735 | < 0.0001 |
| Attention U-Net | 0.8835 ± 0.0087 | 1.4598 ± 0.0760 | 0.4510 ± 0.0291 | 0.0130 | 0.8107 | 0.0737 |
| HippoMamba (Ours) | 0.8861 ± 0.0094 | 1.4533 ± 0.1022 | 0.4430 ± 0.0340 | Reference | Reference | Reference |
| HippoMamba + TTA† (Ours) | 0.8888 ± 0.0095 | 1.4333 ± 0.0925 | 0.4332 ± 0.0357 | n/a | n/a | n/a |

*Dice, HD95 and ASD were computed per subject for the anterior and posterior labels and averaged, then averaged within each fold. All methods used identical fold splits, 64 × 64 × 64 training patches and the checkpoint with the highest validation Dice in each fold. Baselines were trained for up to 200 epochs with early stopping (patience 25) and HippoMamba for 300 epochs. †Eight-flip test-time augmentation applied to HippoMamba only, reported descriptively without a significance test.*

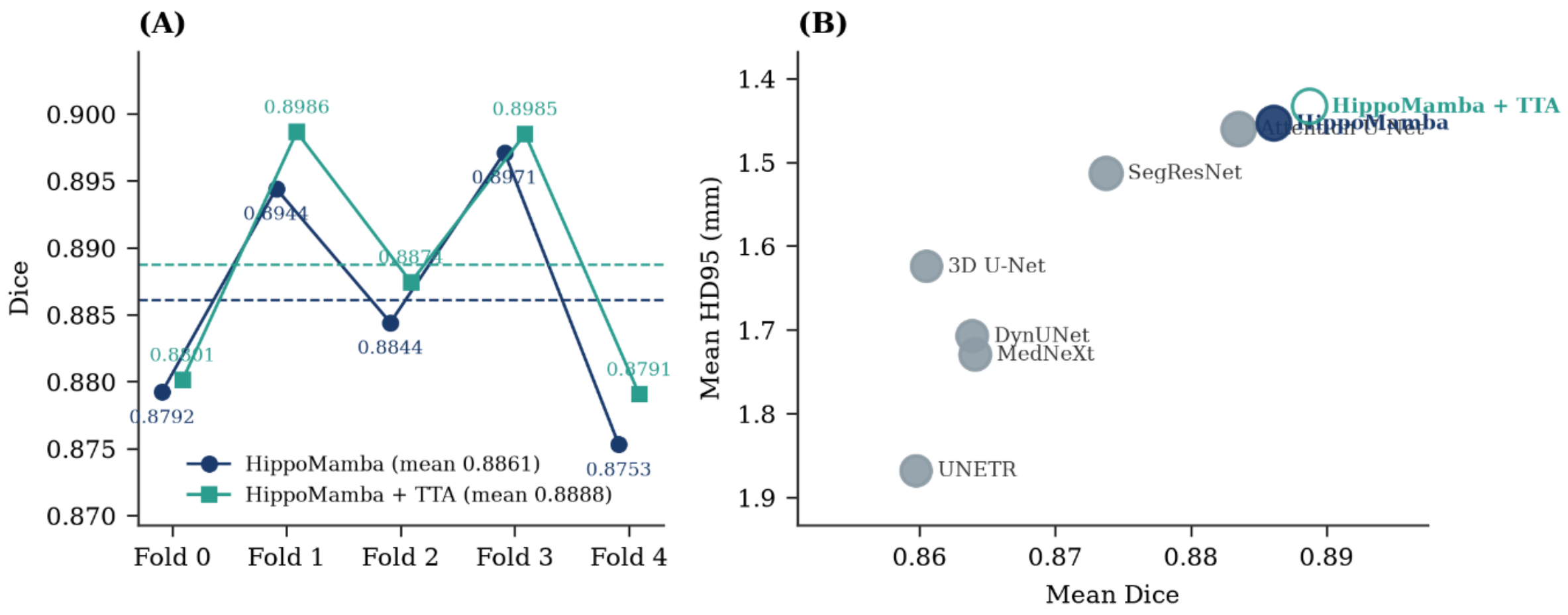

**Fig. 3.** Cross-validation performance on the Medical Segmentation Decathlon Task04 hippocampus dataset. (A) Per-fold Dice of HippoMamba without and with eight-flip test-time augmentation, with values annotated and dashed lines marking the means across folds (0.8861 and 0.8888). (B) Mean Dice against mean HD95 for HippoMamba and the six baselines, with the HD95 axis inverted so that the upper right is favourable and bubble area inversely proportional to ASD, so that larger bubbles indicate smaller surface distance. The open marker shows HippoMamba with test-time augmentation.

Fig. 3A shows the per-fold Dice of HippoMamba, which ranged from 0.8753 to 0.8971 without test-time augmentation (mean 0.8861) and from 0.8791 to 0.8986 with it (mean 0.8888). In Fig. 3B, HippoMamba had the highest mean Dice and the lowest mean HD95 and ASD of the compared methods, and its HD95 and ASD were close to those of Attention U-Net, with differences that were not statistically significant (Table 2).

**Table 3.** Per-fold Dice across the five cross-validation folds

| Method | Fold 0 | Fold 1 | Fold 2 | Fold 3 | Fold 4 | Mean ± SD |
|---|---|---|---|---|---|---|
| UNETR | 0.8563 | 0.8697 | 0.8499 | 0.8684 | 0.8544 | 0.8597 ± 0.0088 |
| 3D U-Net | 0.8607 | 0.8707 | 0.8512 | 0.8670 | 0.8529 | 0.8605 ± 0.0085 |
| DynUNet | 0.8621 | 0.8722 | 0.8593 | 0.8723 | 0.8536 | 0.8639 ± 0.0082 |
| MedNeXt | 0.8622 | 0.8721 | 0.8518 | 0.8765 | 0.8579 | 0.8641 ± 0.0101 |
| SegResNet | 0.8703 | 0.8830 | 0.8646 | 0.8821 | 0.8686 | 0.8737 ± 0.0083 |
| Attention U-Net | 0.8763 | 0.8931 | 0.8795 | 0.8926 | 0.8759 | 0.8835 ± 0.0087 |
| HippoMamba (Ours) | 0.8792 | 0.8944 | 0.8844 | 0.8971 | 0.8753 | 0.8861 ± 0.0094 |
| HippoMamba + TTA† (Ours) | 0.8801 | 0.8986 | 0.8874 | 0.8985 | 0.8791 | 0.8888 ± 0.0095 |

*Values are the mean Dice over the 52 validation subjects of each fold, averaged over the anterior and posterior labels. Mean ± SD is taken across the five fold means. †Eight-flip test-time augmentation applied to HippoMamba only.*

Table 3 reports the Dice of each method in each fold. Without test-time augmentation, HippoMamba had the highest Dice in folds 0 to 3 (0.8792, 0.8944, 0.8844 and 0.8971) and ranked second in fold 4, where Attention U-Net was higher by 0.0006 (0.8759 against 0.8753). With test-time augmentation, HippoMamba had the highest Dice in all five folds. The fold-to-fold standard deviation of HippoMamba (0.0094) was similar to that of the baselines, which ranged from 0.0082 to 0.0101. The variation across folds followed a pattern shared by all methods, each of which reached its highest Dice in fold 1 or 3 and its lowest in fold 2 or 4, indicating that differences between folds mainly reflected the composition of the validation subsets rather than method-specific behaviour.

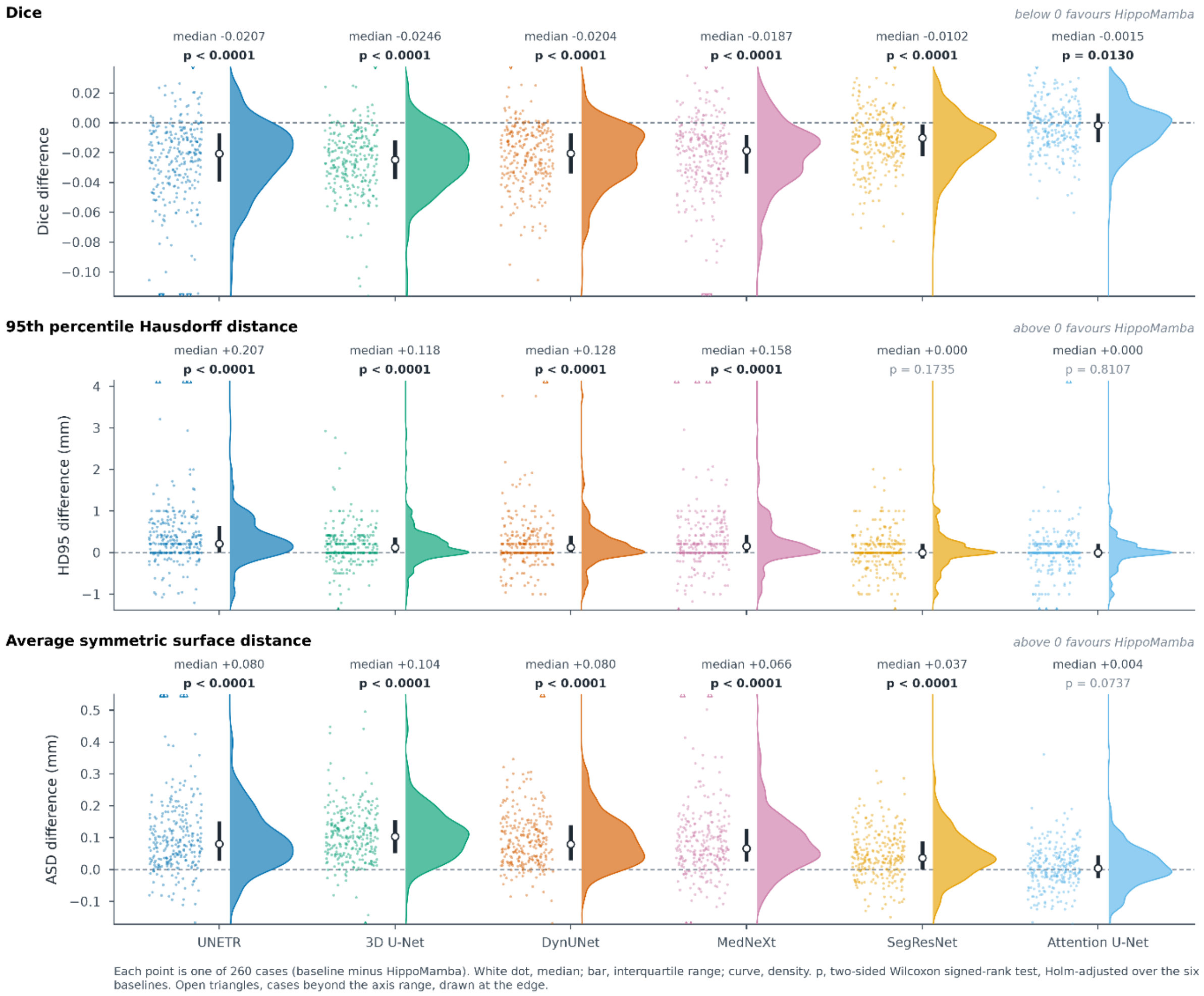

Fig. 4. Per-case paired differences between each baseline and HippoMamba on the Medical Segmentation Decathlon Task04 hippocampus dataset (260 cases, each evaluated by the model of the fold in which it was a validation case). (a) Dice, (b) 95th percentile Hausdorff distance and (c) average symmetric surface distance, each given as the baseline value minus the value of HippoMamba without test-time augmentation, so that negative Dice differences and positive distance differences favour HippoMamba. Each point is one case; the white dot marks the median, the bar the interquartile range and the curve the density. The median difference and the two-sided Wilcoxon signed-rank p-value, Holm-adjusted over the six baselines, are shown above each method. The axes cover the 0.5th to 99.5th percentiles of all differences, and the 12, 14 and 11 cases beyond this range for Dice, HD95 and ASD are drawn at the axis edge as open triangles.

Fig. 4 shows, for each of the 260 subjects, the difference between each baseline and HippoMamba without test-time augmentation, and thus the per-subject distributions underlying the comparisons in Table 2. The median Dice gain of HippoMamba ranged from 0.0246 against 3D U-Net to 0.0015 against Attention U-Net. For Dice and ASD, the median difference favoured HippoMamba against every baseline, and the distributions were shifted away from zero for all baselines except Attention U-Net, whose differences were concentrated close to zero. For HD95, the median difference against SegResNet and Attention U-Net was zero. HD95 takes a small number of discrete values at the 1 mm isotropic voxel size of the dataset, and the two methods had HD95 identical to that of HippoMamba in 82 of 260 subjects (31.5%) and 98 of 260 subjects (37.7%), respectively, consistent with the non-significant tests for these comparisons.

### 5.1.2 Qualitative Analysis

Fig. 5 shows HippoMamba segmentations for three validation cases selected as those closest to the 90th, 50th and 10th percentiles of the per-subject Dice over all 260 cases (hippocampus_229, hippocampus_093 and hippocampus_042, with Dice of 0.9275, 0.8895 and 0.8392). Selecting cases by percentile rather than by visual inspection illustrates upper, typical and lower performance instead of the most favourable examples. For each case, four columns show the input T1-weighted MRI, the manual ground-truth boundary as a cyan contour, the HippoMamba boundary as an orange contour with the Dice inset, and an error map distinguishing false-negative voxels (cyan) from false-positive voxels (orange). In the upper and median cases, the anterior and posterior labels were segmented with similar accuracy (Dice of 0.9319 and 0.9232, and 0.8894 and 0.8896), whereas in the lower case the posterior label was segmented less accurately than the anterior label (0.8120 against 0.8664).

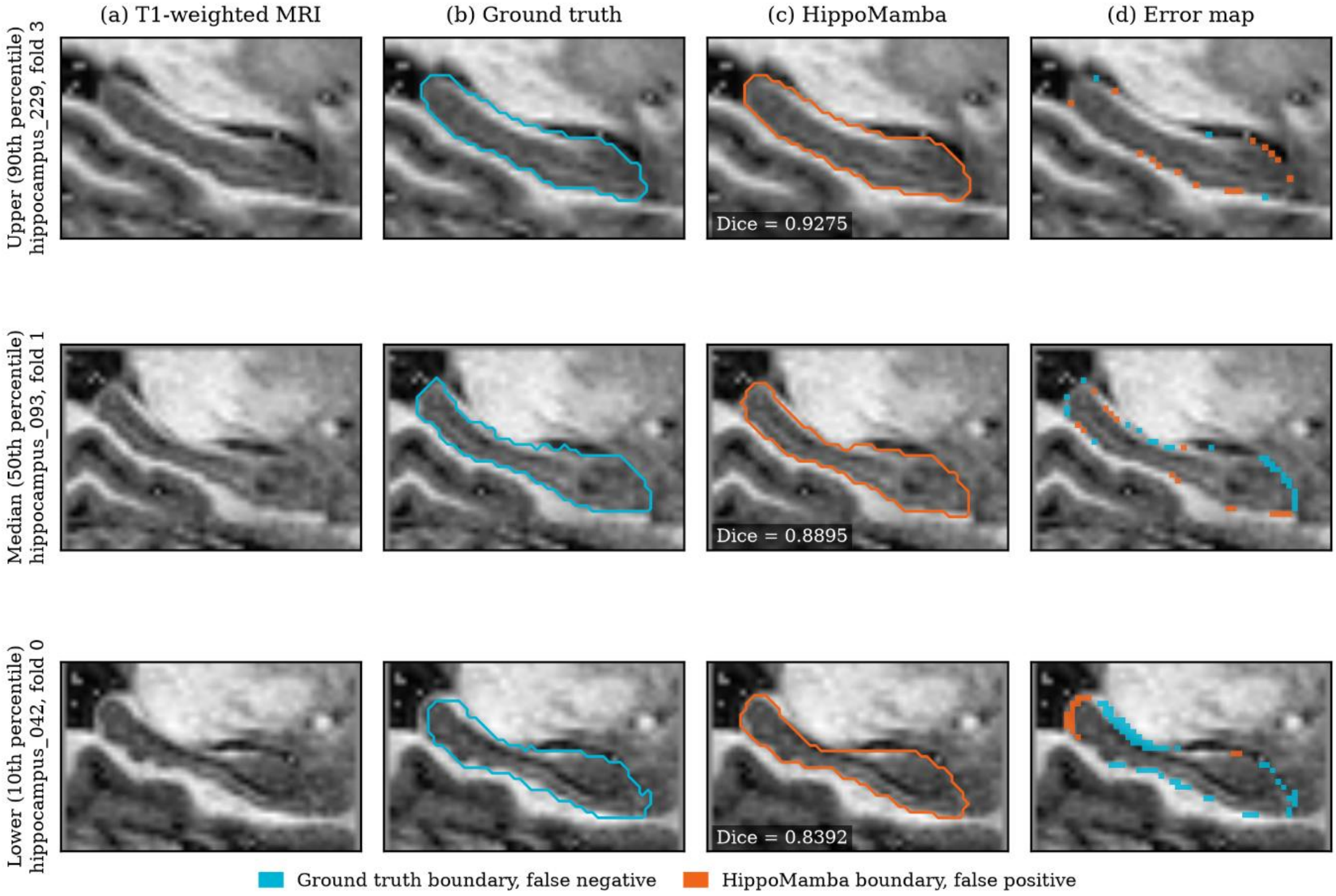

**Fig. 5.** Qualitative segmentation results of HippoMamba on three validation cases of the Medical Segmentation Decathlon Task04 hippocampus dataset, selected as the cases closest to the 90th, 50th and 10th percentiles of the per-subject Dice over all 260 cases (Dice of 0.9275, 0.8895 and 0.8392). Each case was segmented by the model of the fold in which it was a validation case, and the sagittal slice with the largest ground-truth area is shown. Columns show (a) the T1-weighted MRI, (b) the manual ground-truth boundary in cyan, (c) the HippoMamba boundary in orange with the Dice inset, and (d) the error map with false-negative voxels in cyan and false-positive voxels in orange.

## 5.2 External Validation on OASIS-3

To assess generalisation, the five Phase 1 HippoMamba models trained on the Decathlon hippocampus data [28] were applied without retraining or fine tuning to 600 T1-weighted sessions from 150 OASIS-3 participants [29], acquired with voxel sizes of 1.0 × 1.0 × 1.0 mm (361 sessions), 1.2 × 1.05 × 1.05 mm (238 sessions) or 1.0 × 1.02 × 1.02 mm (1 session). One further session was excluded before analysis because it was not a 3D T1-weighted volume. Each scan was affinely registered to the skull-stripped ICBM152 2009a nonlinear MNI template at 1 mm distributed with nilearn [50] by maximising Mattes mutual information [51] within the template brain mask from nine starting positions, and a crop of 35 × 50 × 36 voxels at 1 mm isotropic spacing, matching the median Decathlon image size, was extracted per hemisphere and padded to the 64 × 64 × 64 network input by the Decathlon validation transforms. The five-fold models were combined by averaging their class

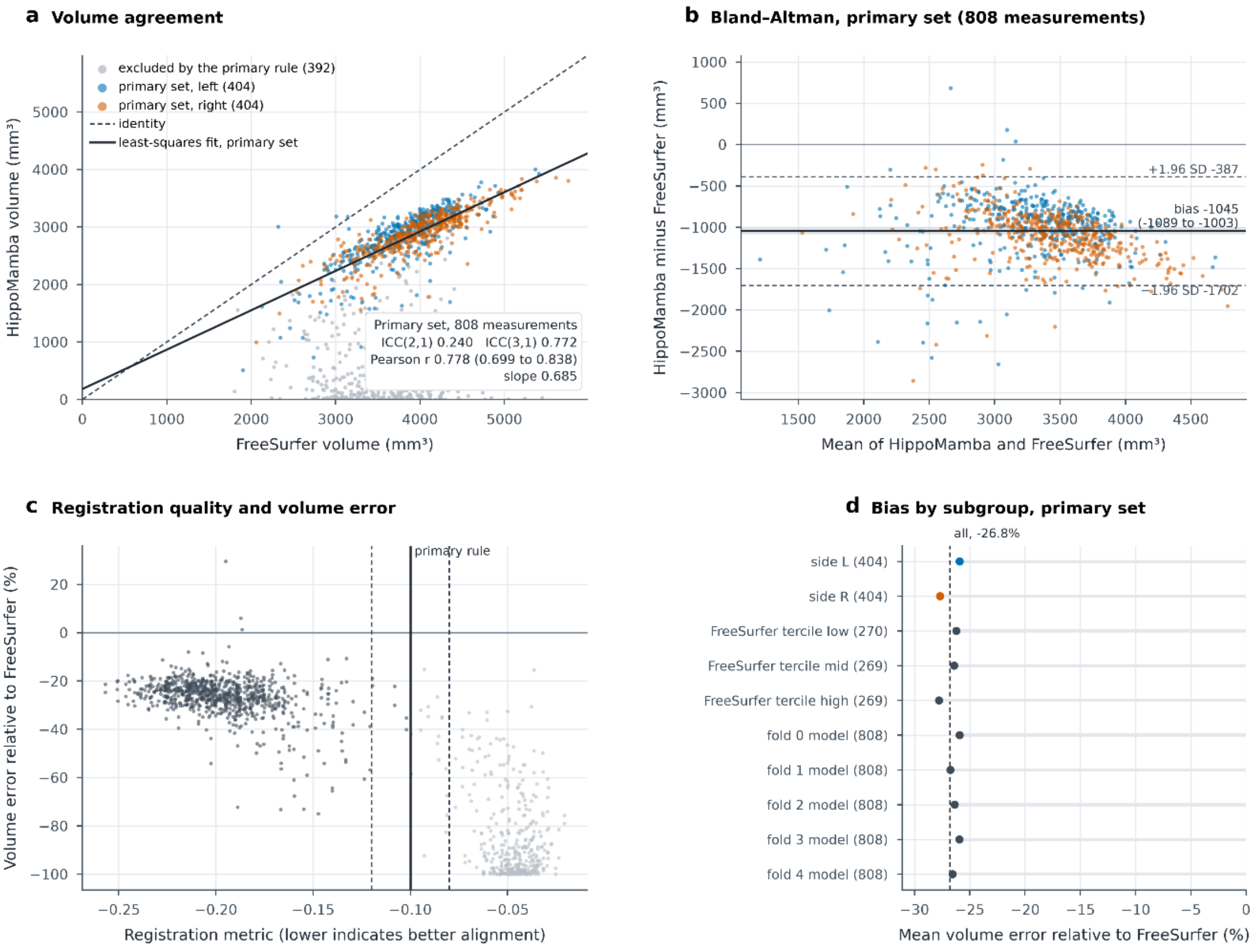

**Fig. 6.** Agreement between HippoMamba and FreeSurfer hippocampal volumes on OASIS-3 (600 sessions of 150 participants, left and right hippocampus, 1,200 measurements). HippoMamba volumes were obtained from the class probabilities averaged over the five fold models trained on the Medical Segmentation Decathlon. (a) HippoMamba against FreeSurfer volume. Measurements from sessions that met the registration criterion (−0.10 or below; 808 measurements from 404 sessions of 138 participants) are coloured by hemisphere, and the 392 excluded measurements are shown in grey. The dashed line is the identity and the solid line the least-squares fit for the analysed sessions. (b) Bland-Altman plot for the analysed sessions, with the mean difference (bias) and its 95% confidence interval (shaded) and the 95% limits of agreement (dashed). (c) Registration metric against the volume error relative to FreeSurfer for all sessions, with the criterion of −0.10 (solid line) and the thresholds of −0.08 and −0.12 (dashed lines). (d) Mean percentage difference from FreeSurfer in the analysed sessions by hemisphere, FreeSurfer volume tercile and fold model, with the overall value as a dashed line.

**Table 4.** Cross-sectional agreement between Hippomamba and Freesurfer hippocampal volumes on OASIS-3

| Registration criterion | Hemispheres | Sessions | Participants | Pearson r (95% CI) | Spearman ρ | ICC (3,1) | ICC (2,1) | MAE (mm³) | MAPE (%) | Bias (mm³) (95% CI) |
|---|---|---|---|---|---|---|---|---|---|---|
| None (all sessions) | 1200 | 600 | 150 | 0.459 (0.386 to 0.526) | 0.605 | 0.352 | 0.136 | 1679 | 45.6 | −1678 (−1785 to −1573) |
| ≤ −0.10 (primary) | 808 | 404 | 138 | 0.778 (0.699 to 0.838) | 0.805 | 0.772 | 0.240 | 1047 | 26.9 | −1045 (−1089 to −1003) |
| ≤ −0.08 | 826 | 413 | 138 | 0.751 (0.653 to 0.823) | 0.787 | 0.748 | 0.243 | 1061 | 27.4 | −1059 (−1106 to −1016) |
| ≤ −0.12 | 802 | 401 | 138 | 0.778 (0.699 to 0.838) | 0.805 | 0.772 | 0.239 | 1047 | 26.9 | −1045 (−1088 to −1003) |

*Predictions are from the five-fold HippoMamba ensemble (64 × 64 × 64 input) applied without retraining. Each session contributes a left and a right hippocampus. The registration criterion is the final value of the registration metric, where lower values indicate better alignment. Bias is predicted minus FreeSurfer volume. MAE, mean absolute error. MAPE, mean absolute percentage error relative to FreeSurfer. ICC (3,1), two-way mixed consistency. ICC (2,1), two-way random absolute agreement. 95% CIs are from a participant-level bootstrap with 2,000 resamples.*

probabilities, and predicted volumes were counted on the padded grid and compared with the FreeSurfer hippocampal volumes distributed with the dataset [6], [27]. In 13 of the 1,200 hemisphere measurements, all outside the analysed sessions, the models also labelled voxels in the padding (at most 217 mm³). Sessions with a final registration metric of −0.10 or below (404 of 600 sessions from 138 participants), a criterion fixed before evaluation of the reported models, were analysed, and thresholds of −0.08 and −0.12 gave the same conclusions (Pearson r from 0.751 to 0.778, Table 4). Agreement was summarised with Pearson and Spearman correlation, ICC (3,1) for consistency and ICC (2,1) for absolute agreement [47], [48], and the Bland-Altman bias and 95% limits of agreement [49], with 95% confidence intervals from a participant-level bootstrap of 2,000 resamples. As shown in Table 4, the ensemble ranked hippocampal volumes consistently with FreeSurfer (Pearson r = 0.778, 95% CI 0.699 to 0.838; Spearman ρ = 0.805; ICC (3,1) = 0.772), while absolute volumes were smaller (mean bias −1,045 mm³, 95% CI −1,089 to −1,003; 95% limits of agreement −1,702 to −387 mm³; ICC (2,1) = 0.240). The difference was proportional (predicted volume ≈ 0.69 × FreeSurfer volume + 180 mm³), and the percentage difference was similar across volume terciles, hemispheres and fold models (−25.9% to −27.8%). Fig. 6 shows the individual measurements (Fig. 6a), the Bland-Altman agreement (Fig. 6b), the volume error against the registration metric for all sessions (Fig. 6c) and the bias in each subgroup (Fig. 6d).

For longitudinal change, annual rates (least-squares slope divided by mean volume) were computed for participants with at least two sessions (Table 5). In the primary analysis (238 participant-hemispheres from 119 participants, median span 6.0 years), the correlation between predicted and FreeSurfer rates was r = 0.569 (95% CI 0.263 to 0.570; Spearman ρ = 0.389; mean absolute difference 4.95 %/yr), but this estimate was strongly influenced by 16 participant-hemispheres from 8 participants with follow-up of 0.01 to 1.94 years, for which annualised FreeSurfer rates reached 728 %/yr (mean 66.3 %/yr; Fig. 7a). In the pre-specified sensitivity analysis restricted to follow-up of at least two years (222 participant-hemispheres

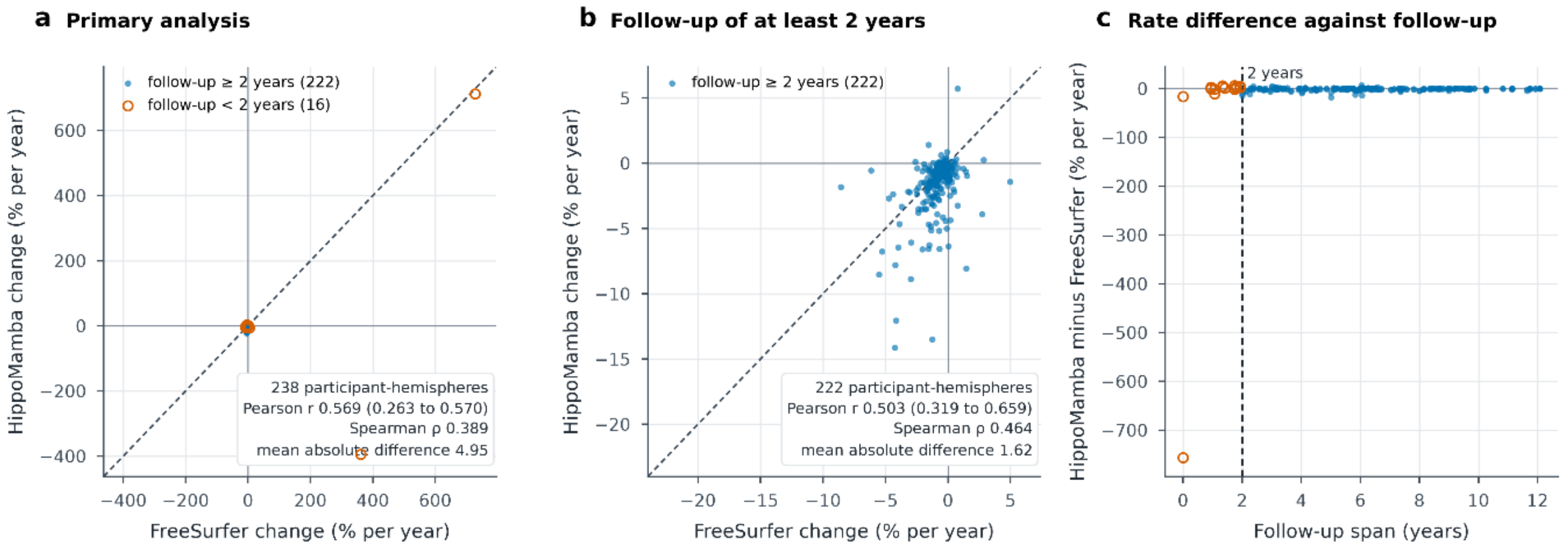

**Fig. 7.** Annualised hippocampal volume change of HippoMamba and FreeSurfer on OASIS-3 for participant-hemispheres with at least two sessions that met the registration criterion. (a) Primary analysis (238 participant-hemispheres from 119 participants), with the 16 participant-hemispheres from 8 participants whose follow-up was shorter than 2 years (0.01 to 1.94 years) shown as open circles. (b) Sensitivity analysis restricted to follow-up of at least 2 years (222 participant-hemispheres from 111 participants). The dashed line is the identity, and each box gives the Pearson correlation with its 95% participant-level bootstrap confidence interval, the Spearman correlation and the mean absolute difference in percent per year. (c) Difference between the HippoMamba and FreeSurfer rates against the follow-up span, with the 2-year threshold as a dashed line. Annualised change is the least-squares slope of volume against time divided by the participant-hemisphere mean volume, multiplied by 100.

**Table 5.** Agreement between Hippomamba and Freesurfer annualised hippocampal volume change on OASIS-3

| Analysis | Participant-hemispheres | Participants | Median span (years) | HippoMamba (%/yr) | FreeSurfer (%/yr) | Pearson r (95% CI) | Spearman ρ | MAE (%/yr) |
|---|---|---|---|---|---|---|---|---|
| No registration criterion | 300 | 150 | 7.5 | −10.58 | −0.94 | 0.081 (−0.040 to 0.217) | 0.187 | 12.08 |
| Registration ≤ −0.10 (primary) | 238 | 119 | 6.0 | −0.65 | +3.54 | 0.569 (0.263 to 0.570) | 0.389 | 4.95 |
| Registration ≤ −0.10 and span ≥ 2 years (sensitivity) | 222 | 111 | 6.1 | −2.05 | −0.99 | 0.503 (0.319 to 0.659) | 0.464 | 1.63 |

*Annualised change is the least-squares slope of volume against time divided by the participant-hemisphere mean volume, multiplied by 100. Values are means over participant-hemispheres. MAE is the mean absolute difference between HippoMamba and FreeSurfer rates. In the primary analysis the bootstrap distribution of Pearson r was asymmetric and its interval does not centre on the point estimate, reflecting the influence of 16 participant-hemispheres with follow-up shorter than two years. 95% CIs are from a participant-level bootstrap with 2,000 resamples.*

from 111 participants, median span 6.1 years), rates were −2.05 %/yr for HippoMamba and −0.99 %/yr for FreeSurfer (Pearson r = 0.503, 95% CI 0.319 to 0.659; Spearman ρ = 0.464), with a mean absolute difference of 1.63 %/yr (Fig. 7b). Fig. 7c shows the difference between the two rates against the follow-up span. Several limitations apply. The reference was an automated FreeSurfer segmentation and not a manual one, and FreeSurfer has been reported to give larger hippocampal volumes than manual tracing in paediatric samples [52], so part of the volume offset may reflect differences in labelling conventions, which could not be assessed without an overlap metric. Roughly one third of the sessions fell outside the registration criterion, and without any criterion the agreement was lower (r = 0.459, ICC (3,1) = 0.352), so the results describe sessions with adequate registration. Results were also not stratified by scanner or field strength.

## 6. Discussion

HippoMamba combines a volumetric convolutional segmentation network with a Mamba selective state space block at the bottleneck, designed to process one or more MRI visits of a subject within a single model. The results presented in Section 5 show that this design achieves the highest Dice among the compared segmentation methods under identical folds and patch size, and that, when applied without retraining to an independent multi-scanner cohort, it produces hippocampal volumes and volume changes that are consistent with an established reference method.

### 6.1 Segmentation performance

The Phase 1 segmentation results establish HippoMamba as a competitive method on the MSD Task04 Hippocampus benchmark, with a five-fold cross-validation mean Dice of 0.8861 without test-time augmentation, the highest among the compared methods, and per-subject Dice significantly higher than that of each of the six baselines. The margin over the strongest baseline, Attention U-Net, was small (median paired difference 0.0015, $p = 0.0130$), and the boundary metrics of the two methods did not differ significantly. Attention U-Net [31] was clearly stronger than the other baselines, which places the benchmark close to its performance ceiling for architectures of this kind and indicates that differences of this magnitude should be interpreted with care. Two aspects of the comparison are relevant to this interpretation. First, the training settings were not identical across methods, as HippoMamba was trained for 300 epochs with a batch size of one, whereas the baselines were trained with early stopping and, in four cases, with a constant learning rate and simpler preprocessing (Section 4.3.1). Second, the contribution of the Mamba block to segmentation was not isolated in this study, and for single scans the temporal encoder operates on a sequence of length one, so the segmentation results should be attributed to the network as a whole rather than to state space modeling specifically. Test-time augmentation increased the mean Dice of HippoMamba by a further 0.0027, which is consistent with the usual benefit of flip ensembling for small structures.

### 6.2 External validation and serial volumetry

Applied without retraining to 600 sessions from 150 OASIS-3 participants [29], the five-fold ensemble produced hippocampal volumes that ranked consistently with FreeSurfer (Pearson $r = 0.778$, ICC (3,1) = 0.772 under the pre-specified registration criterion), and the conclusions were unchanged under alternative thresholds. Absolute agreement was lower (ICC (2,1) = 0.240), because predicted volumes were systematically about 27% smaller than FreeSurfer volumes, with a proportional relationship that was similar across hemispheres, volume terciles, and fold models. A difference of this kind is consistent with differing anatomical definitions between the manual Decathlon labels used for training and the automated FreeSurfer segmentation, and with reports that FreeSurfer yields larger hippocampal volumes than manual tracing [52], although it could not be attributed to a specific cause without an overlap reference. For serial volumetry, annualized volume change derived from per-visit segmentation was moderately correlated with that of FreeSurfer in participants followed for at least two years ($r = 0.503$), although the mean HippoMamba rate was about twice as large (−2.05 against −0.99 %/yr), whereas the primary longitudinal estimate was strongly influenced by short follow-up intervals, for which annualized rates are dominated by measurement noise in both methods. These findings indicate that HippoMamba volumes are suitable for ranking hippocampal volume within a cohort, while absolute volumes and rates of change should be interpreted relative to the reference method.

### 6.3 Mamba for longitudinal neuroimaging

The choice of Mamba as the temporal modeling component of HippoMamba is motivated by several architectural considerations relevant to longitudinal neuroimaging. The number of serial MRI visits per subject in typical clinical cohorts is small, ranging from 2 to 5 in the ADNI-3 data analyzed here, which places the application in the short-sequence regime where the selective state update mechanism is more relevant than the linear complexity advantage of Mamba over transformers [19]. The selective scan allows the model to gate the influence of each visit on the accumulated hidden state, which is conceptually aligned with the observation that atrophy trajectories in AD are non-linear and that the informativeness of individual visits depends on disease stage [35]. The recurrent formulation in Equations (7) and (8) also handles sequences of variable length without padding-aware attention masks. Mamba-based models have shown competitive performance in 3D medical image segmentation [21], [23], [25] and in pathology analysis requiring long-sequence modeling [26], but the use of the sequence dimension for serial patient visits differs from these applications. In the present study, the longitudinal predictions did not add information beyond the scan schedule (Section 6.4), so whether temporal state space modeling provides a useful inductive bias for serial MRI remains an open question.

### 6.4 Limitations and future work

The longitudinal module was evaluated with subject-level five-fold cross-validation in 182 ADNI-3 participants with at least two visits, so that each participant was predicted by a model that had not seen them during training or checkpoint selection. Predicted and FreeSurfer fractional changes in hippocampal volume were positively correlated (Pearson $r = 0.324$, 95% CI 0.184 to 0.450). A linear model using only follow-up duration and the number of visits reached a higher correlation under the same folds ($r = 0.370$, 95% CI 0.232 to 0.491), with a difference between the two correlations of −0.046 (95% CI −0.094 to −0.001, $p = 0.048$), and after adjustment for these two variables the partial correlation between predicted and FreeSurfer change was −0.179 (95% CI −0.333 to −0.023). The longitudinal predictions therefore did not add information about hippocampal change beyond the scan schedule, and differences in predicted annualized change between diagnostic groups largely reflected differences in follow-up duration. Longitudinal volumetry derived from per-visit segmentation, as in the OASIS-3 analysis, provides a more direct measure, and conditioning the temporal module on acquisition intervals in larger longitudinal cohorts, such as additional ADNI phases or AIBL, is a direction for future work. The reference volumes for both ADNI-3 and OASIS-3 were obtained with cross-sectional FreeSurfer processing rather than manual tracing, which limits the precision of both the training targets and the evaluation of volume change, particularly over short intervals.

Segmentation relies on hippocampus-centered inputs, which are provided by the Decathlon data and were obtained for OASIS-3 through registration to a template, and about one third of the OASIS-3 sessions did not meet the registration criterion, so the external validation describes sessions with adequate registration. Integration of a learned localization step or a whole-brain segmentation stage would reduce this dependency. The cross-validation results of all methods are expected to be slightly optimistic, because the checkpoint was selected on the same validation subjects on which performance is reported, and the external validation was not stratified by scanner or field strength. During HippoMamba training, gradient-norm clipping was applied to the gradients scaled by the mixed-precision loss scaler, which reduces the effective magnitude of parameter updates, and the effect of this choice on the reported performance was not examined. Extension to subfield-level quantification would provide richer spatial information about hippocampal subregions known to exhibit differential vulnerability in AD [4]. Finally, recent Mamba-based segmentation methods such as U-Mamba [20], VM-UNet [22], SegMamba [21], and SegMamba-V2 [25] were not included in the comparison, and a comparison with these methods, together with an ablation of the Mamba block for single-timepoint segmentation, is left to future work.

## 7. Conclusion

This paper presented HippoMamba, a network for hippocampal segmentation and serial volumetry in Alzheimer's disease that combines a three-dimensional convolutional encoder and decoder with a Mamba selective state space block at the bottleneck, designed to process one or more MRI visits of a subject within a single model. A two-phase training strategy decouples single-timepoint segmentation learning on an annotated benchmark from longitudinal modeling on a clinical cohort, so that the framework can be trained without longitudinal voxel-wise labels. On the Medical Segmentation Decathlon Task04 benchmark, under identical folds and patch size and a single evaluation procedure, HippoMamba achieved the highest mean Dice among six convolutional and transformer-based methods, with per-subject Dice significantly higher than that of each method, and the lowest mean Hausdorff distance and average surface distance, although the boundary metrics did not differ significantly from those of the strongest baseline. Applied without retraining to an independent multi-scanner cohort, HippoMamba produced hippocampal volumes that ranked consistently with FreeSurfer but were systematically about 27% smaller, and annualized volume change derived from per-visit segmentation was moderately correlated with that of FreeSurfer in participants followed for at least two years, with a larger mean rate. These results support the use of HippoMamba for hippocampal segmentation and for ranking hippocampal volume within a cohort, with absolute volumes and rates of change interpreted relative to the reference method. Conditioning the temporal module on acquisition intervals and evaluating it in larger longitudinal cohorts are directions for future work toward direct prediction of hippocampal atrophy from serial MRI.

## Ethics Statement

This study used three publicly available, de-identified datasets. ADNI data collection procedures were approved by the institutional review boards at each participating site, and written informed consent was obtained from all participants or their authorized representatives at the time of original data collection. OASIS-3 data were collected under protocols approved by the Institutional Review Board of Washington University in St. Louis with written informed consent from all participants, and were obtained under the OASIS data use agreement. The Medical Segmentation Decathlon hippocampus data are publicly released for research use. No additional ethical approval was required for the present study, as it involved secondary analysis of fully de-identified, publicly available imaging and clinical data obtained under the respective data use agreements.

## Acknowledgements

Data used in preparation of this article were obtained from the Alzheimer's Disease Neuroimaging Initiative (ADNI) database (adni.loni.usc.edu). A complete listing of ADNI investigators can be found at http://adni.loni.usc.edu/wp-content/uploads/how_to_apply/ADNI_Acknowledgement_List.pdf. Data were provided in part by OASIS-3: Longitudinal Multimodal Neuroimaging: Principal Investigators: T. Benzinger, D. Marcus, J. Morris; NIH P30 AG066444, P50 AG00561, P30 NS09857781, P01 AG026276, P01 AG003991, R01 AG043434, UL1 TR000448, R01 EB009352. The authors thank the organizers of the Medical Segmentation Decathlon for making the hippocampus data publicly available.